\pdfoutput=1 

\documentclass{article}

\PassOptionsToPackage{numbers}{natbib}

\usepackage[final]{neurips_data_2021}

\usepackage[utf8]{inputenc} 
\usepackage[T1]{fontenc}    

\usepackage{url}            
\usepackage{booktabs}       
\usepackage{amsfonts}       
\usepackage{nicefrac}       
\usepackage{microtype}      
\usepackage[table]{xcolor}
\usepackage{multirow}

\newcommand{\eg}{{\em e.g.,~}}
\newcommand{\ie}{{\em i.e.,~}}

\newcommand{\etal}{{\em ~et al.}}

\usepackage[disable,bordercolor=white,backgroundcolor=gray!30,linecolor=black,colorinlistoftodos]{todonotes}

\usepackage[ruled,vlined,linesnumbered]{algorithm2e}

\SetCommentSty{mycommfont}

\newcommand{\deleted}[1]{\todo[inline,color=gray!40]{{\bf Deleted:} #1}}
\newcommand{\substituted}[1]{\todo[inline,color=gray!40]{{\bf Substituted:} #1}}
\newcommand{\comment}[1]{\todo[inline,color=lightgray!40]{{\bf Comment:} #1}}

\usepackage{amsmath}
\usepackage{amssymb}
\usepackage{bbm}
\usepackage{amsthm}

\usepackage{makecell}

\usepackage{placeins}

\usepackage{capt-of}
\usepackage{varwidth}

\newcommand{\haozhe}[1]{\todo[inline,color=green!40]{#1 -- Haozhe}}
\newcommand{\isabelle}[1]{\todo[inline,color=orange!40]{#1 -- Isabelle}}

\usepackage{adjustbox}

\title{OmniPrint: \\ A Configurable Printed Character Synthesizer}

\author{Haozhe Sun\(^*\), Wei-Wei Tu\(^{\#+}\), Isabelle Guyon\(^{*+}\) \\
  \(*\) LISN (CNRS/INRIA) Université Paris-Saclay, France \\
  \# 4Paradigm Inc, Beijing, China \\
  + ChaLearn, California, USA \\
  \texttt{omniprint@chalearn.org }}

\begin{document}

\maketitle



\vspace{-0.7cm}
\begin{abstract}
  We introduce OmniPrint, a synthetic data generator of isolated printed characters, geared toward machine learning research. It draws inspiration from famous datasets such as MNIST, SVHN and Omniglot, but offers the capability of generating a wide variety of printed characters from various languages, fonts and styles, with customized distortions. We include 935 fonts from 27 scripts and many types of distortions. 
  As a proof of concept, we show various use cases, including an example of meta-learning dataset designed for the upcoming MetaDL NeurIPS 2021 competition. OmniPrint is available at \href{https://github.com/SunHaozhe/OmniPrint}{https://github.com/SunHaozhe/OmniPrint}.
  
  
  
\end{abstract}

\deleted{Use cases include: image classification benchmarks, data augmentation, calibration, bias detection/compensation, meta-learning and transfer learning,
  modular learning from decomposable/separable problems, recognition of printed characters in the wild, and creation of captchas.}
 \comment{We will provide a few more use cases for illustration purposes in the discussion section, but less developed. We are de-emphasizing possible OCR applications, which are not illustrated in the paper. Added text is highlighted in yellow.}


\begin{figure}[hb]
\vspace{-0.3cm}
    \centering
    \includegraphics[width=0.8\linewidth]{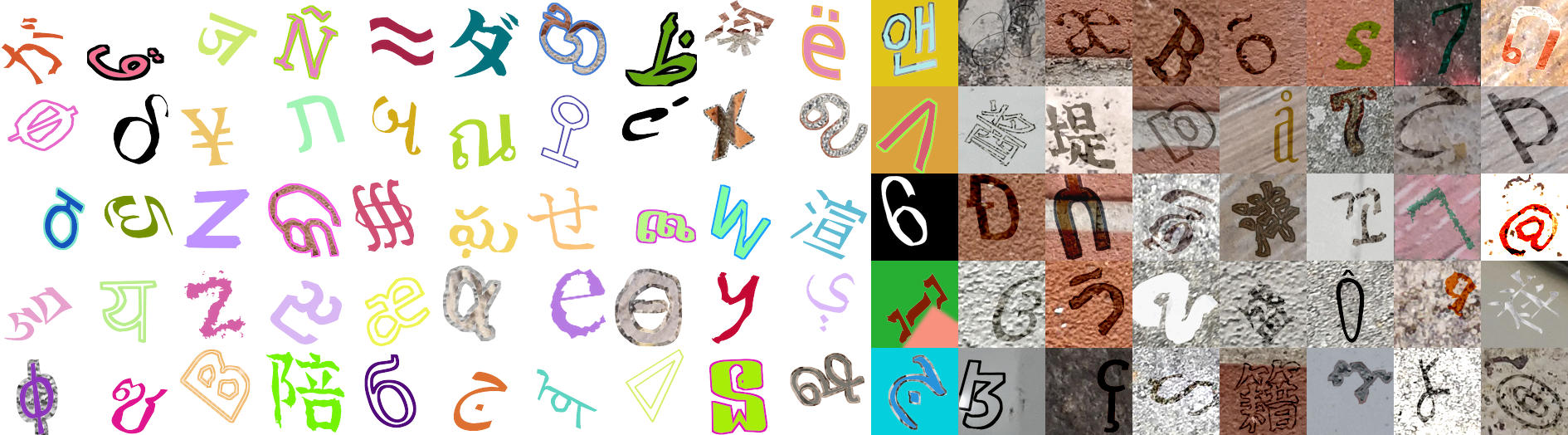}
    \caption{{\bf Examples of characters generated by OmniPrint.}}
\label{fig:samples}
    \vspace{-0.3cm}
\end{figure}

\section{Introduction and motivation}
\label{sec:introduction}

Benchmarks and shared datasets have been fostering progress in Machine learning (ML)~\cite{2014arXiv1409.0575R, Krizhevsky09learningmultiple, WelinderEtal2010, LFWTech}. One of the most popular benchmarks is MNIST~\cite{lecun2010mnist}, which is used all over the world in tutorials, textbooks, and classes. Many variants of MNIST have been created~\cite{cohen_afshar_tapson_schaik_2017, clanuwat2018deep, mu2019mnist, DBLP:journals/corr/SrivastavaMS15, ganinDomainAdversarialTrainingNeural2016, yadavColdCaseLost2019}, including one created recently, which inspired us: Omniglot~\cite{lakeHumanlevelConceptLearning2015}. This dataset includes characters from many different scripts. Among machine learning techniques using such benchmark datasets, Deep Learning techniques are known to be very data hungry. Thus, while there is an increasing number of available datasets, there is a need for larger ones. But, collecting and labeling data is time consuming and expensive, and systematically varying environment conditions is difficult and necessarily limited. Therefore, resorting to artificially generated data is useful to drive fundamental research in ML. This motivated us to create OmniPrint, as an extension to Omniglot, geared to the generation of an unlimited amount of printed characters.

\deleted{There exist also a number of Optical Character Recognition datasets (OCR), including SVHN~\cite{netzerReadingDigitsNatural2011}, aiming at pushing the state-of-the-art in OCR "in the wild", with applications including self-driving car. Modern OCR techniques often involve Deep Learning techniques, which are known to be data hungry. While there is an increasing number of available datasets, collecting and labeling data is time consuming and expensive. Additionally, systematically varying environment condition is difficult and necessarily limited. Thus resorting to artificially generated data is useful both to drive fundamental research in ML and to push the state-of-the art in OCR.}

Of all ML problems, we direct our attention to classification and regression problems in which a vector \({\bf y}\) (discrete or continuous labels) must be predicted from a real-valued input vector \({\bf x}\) of observations (in the case of OmniPrint, an image of a printed character). Additionally, data are plagued by nuisance variables \mbox{\({\bf z}\)}, another vector of discrete or continuous labels, called \mbox{{\em metadata}} or \mbox{{\em covariates}}. In the problem at hand, \mbox{\({\bf z}\)} may include various character distortions, such as shear, rotation, line width variations, and changes in background. Using capital letters for random variable and lowercase for their associated realizations, a data generating process supported by OmniPrint  consists in three steps:
\begin{eqnarray}
{\mathbf z} & \sim &  \mathbb{P}({\bf Z}) \\
{\mathbf y} & \sim &  \mathbb{P}({\bf Y | Z}) \\
{\mathbf x} & \sim &  \mathbb{P}({\bf X | Z, Y}) 
\end{eqnarray}
Oftentimes, \mbox{\({\bf Z}\)} and \mbox{\({\bf Y}\)} are independent, so \mbox{\(\mathbb{P}({\bf Y | Z}) = \mathbb{P}({\bf Y})\)}. This type of data generating process is encountered in many domains such as image, video, sound, and text applications (in which objects or concepts are target values \mbox{\({\bf y}\)} to be predicted from percepts \mbox{\({\bf x}\))};  medical diagnoses of genetic disease (for which \mbox{\({\bf x}\)} is a phenotype and \mbox{\({\bf y}\)} a genotype); analytical chemistry (for which \mbox{\({\bf x}\)} may be chromatograms, mass spectra, or other instrument measurements, and \mbox{\({\bf y}\)} compounds to be identified), etc. Thus, we anticipate that progress made using OmniPrint to benchmark machine learning systems should also foster progress in these other domains.

\deleted{Data generative processes underlying classification problems can be thought of as a joint distribution \(\mathbb{P}({\bf X}, {\bf Y})\) between observed variables \({\bf x}\), which are realizations of a random vector \({\bf X}\) (\eg images), and labels to be predicted \({\bf y}\), which are realizations of a random vector \({\bf Y}\) (\eg image labels) \cite{Vapnik1998}. Several causal mechanisms~\cite{Peters17} can explain such joint distribution, including \({\bf Y} = f({\bf X}, {\bf Z})\) or \({\bf X} = f({\bf Y}, {\bf Z})\), where \(f\) ia a functions, and \({\bf Z}\) is a random vector of context or latent variables, which can be thought of as "bias" or "noise", depending on the situation. More complex causal relationships arise \eg when subsets of variables of \({\bf Z}\) are either causes or consequences of  \({\bf Y}\). For OCR problems, we are in the case \({\bf X} = f({\bf Y}, {\bf Z})\), where \({\bf Y}\) is the class label (character), \({\bf Z}\) encompasses font, style, distortions, foreground, background, noise, etc., and \({\bf X}\) is the generated image. This should allow us to foster progress in {\bf a wide variety of problems, whose generative processes are similar}, such as image, video, sound, and text applications (in which objects or concepts are target values \({\bf y}\) to be predicted from percepts \({\bf x}\));  medical diagnoses of genetic disease (for which \({\bf x}\) is a phenotype and \({\bf y}\) a genotype); analytical chemistry (for which \({\bf x}\) may be chromatograms, mass spectra, or other instrument measurements, and \({\bf y}\) compounds to be identified), etc. To complete the data generative process, one may assume that \({\bf Y}\) and  \({\bf Z}\) are independent random variables drawn from \(\mathbb{P}({\bf Y})\) and  \(\mathbb{P}({\bf Z})\). However in some cases, \({\bf Y}\) and  \({\bf Z}\) are not independent \eg classifying characters within an alphabet; thus an alphabet \({\bf z}\) is drawn first from \(\mathbb{P}({\bf Z})\), then a character class \({\bf y}\) is drawn from \(\mathbb{P}({\bf Y} | {\bf Z})\). Finally, OmniPrint is not strictly limited to classification problems: even though predicting the character class is the natural application, by exchanging the role of the class labels \({\bf y}\) and some of the components of \({\bf z}\) (font, distortions, noises, etc.) one could create regression problems.}

\deleted{"by focusing on OCR, }


Casting the problem in such a generic way should allow researchers to target a variety of ML research topics. Indeed, character images provide excellent benchmarks for machine learning problems because of their relative simplicity, their visual nature, while opening the door to high-impact real-life applications. However, our survey of available resources (Section~\ref{sec:related}) revealed that no publicly available data synthesizer fully suits our purposes: generating realistic quality images \({\bf x}\) of small sizes (to allow fast experimentation) for a wide variety of characters \({\bf y}\) (to study extreme number of classes), and wide variety of conditions parameterized by \({\bf z}\) (to study invariance to realistic  distortions). A conjunction of technical features is required to meet our specifications: pre-rasterization manipulation of anchor points; post-rasterization distortions; natural background and seamless blending; foreground filling; anti-aliasing rendering; importing new fonts and styles.

\deleted{including  data augmentation~\cite{simardBestPracticesConvolutional2003}, calibration, bias detection/compensation, meta-learning~\cite{hospedalesMetaLearningNeuralNetworks2020} and transfer learning~\cite{panSurveyTransferLearning2010} (and their variants: co-covariate shift, multi-task learning, domain adaptation, few-shot learning, combinatorial generalization, zero-shot learning); modular learning from decomposable/separable problems~\cite{aletModularMetalearning2019, lakeCompositionalGeneralizationMeta2019}}

\deleted{from "OCR problems provide ideal benchmarks for such problems because of their relative simplicity, their visual nature, and the possibility of generating synthetic data, while opening the door to high-impact application scenarios of {\em recognition of printed characters in the wild}" to "In particular, character images provide ideal benchmarks for Machine Learning problems because of their relative simplicity, their visual nature, and the possibility of generating synthetic data, while opening the door to high-impact real-life applications}



Modern fonts (\eg TrueType or OpenType) are made of straight line segments and quadratic Bézier curves, connecting anchor points. Thus it is easy to modify characters by moving anchor points. This allows users to perform vectors-space pre-rasterization geometric transforms (rotation, shear, etc.) as well as distortions (\eg modifying the length of ascenders of descenders), without incurring aberrations due to aliasing, when transformations are done in pixel space (post-rasterization). The closest software that we found fitting our needs is \href{https://github.com/Belval/TextRecognitionDataGenerator}{"Text Recognition Data Generator"}~\cite{BelvalTe73:online} (under MIT license), which we used as a basis to develop OmniPrint. While keeping the original software architecture, we substituted individual components to fit our needs. 
Our contributions include:
(1) Implementing many {\bf new transformations and styles}, \eg elastic distortions, natural background, foreground filling, etc.; (2) Manually selecting characters from the Unicode standard to form alphabets from {\bf more than 20 languages around the world}, further grouped into partitions, to facilitate creating meta-learning tasks; (3) Carefully {\bf identifying fonts}, which suit these characters; 
(4) Replacing character rendering by a low-level FreeType font rasterization engine~\cite{TheFreeT60:online}, which enables {\bf direct manipulation of anchor points}; (5) Adding {\bf anti-aliasing rendering}; (6) Implementing and optimizing utility code to facilitate {\bf dataset formatting}; (7) Providing a meta-learning {\bf use case} with a sample dataset.
To our knowledge, OmniPrint is the first text image synthesizer geared toward ML research, supporting pre-rasterization transforms. 
This allows Omniprint to imitate handwritten characters, to some degree. 

\substituted{ "OCR" by "text image"}

\deleted{While our focus is on generating isolated characters for ML research, OmniPrint can also generate text for OCR research.}




\section{Related work}


\label{sec:related}


While our focus is on generating isolated characters for ML research, related work is found in OCR research and briefly reviewed here. OCR problems include {\bf recognition of text from scanned documents} and {\bf recognition of characters "in the wild"} from pictures of natural scenes:

{\bf - OCR from scanned documents} is a well developed field. There are many systems performing very well on this problem~\cite{netzerReadingDigitsNatural2011, jaderberg_synthetic_2014, chen_text_2020}. Fueling this research, many authors have addressed the problem of generating artificial or semi-artificial degraded text since the early 90's~\cite{kanungo_global_1993}. More recently, Kieu\etal~\cite{kieu_character_2012} simulate the degradation of aging document and the printing/writing process, such as dark specks near characters or ink discontinuities, and Kieu\etal~\cite{kieu_efficient_2013} extend this work by facilitating the parameterization. Liang\etal~\cite{liang_geometric_2008} generalize the perspective distortion model of Kanungo\etal~\cite{kanungo_global_1993} by modeling thick and bound documents as developable surfaces. Kieu\etal~\cite{kieu_semi-synthetic_2013} present a 3D model for reproducing geometric distortions such as folds, torns or convexo-concaves of the paper sheet. Besides printed text, handwritten text synthesis has also been investigated, \eg~\cite{gravesGeneratingSequencesRecurrent2014}. 

{\bf - Text in the wild, or scene text}, refer to text captured in natural environments, such as sign boards, street signs, etc. yielding larger variability in size, layout, background, and imaging conditions. Contrary to OCR in scanned documents, scene text analysis remains challenging. Furthermore, the size of existing real scene text datasets is still small compared to the demand of deep learning models. Thus, synthetic data generation is an active field of research~\cite{chen_text_2020}. Early works did not use deep learning for image synthesis. They relied on font manipulation and traditional image processing techniques, synthetic images are typically rendered through several steps including font rendering, coloring, perspective transformation, background blending, etc.~\cite{de_campos_character_2009, wang_end--end_2012, jaderberg_synthetic_2014}. In recent years, text image synthesis involving deep learning has  generated impressive and photo-realistic text in complex natural environments~\cite{gregorDRAWRecurrentNeural2015,netzerReadingDigitsNatural2011,Gupta16,bustaE2EMLTUnconstrainedEndtoEnd2018,zhan_verisimilar_2018,long_unrealtext_2020,zhan_spatial_2019, yang_controllable_2019, zhan_ga-dan_2019,wu_editing_2019, yang_swaptext_2020}. We surveyed the Internet for open-source text generation engines. The most popular ones include SynthText~\cite{Gupta16}, UnrealText~\cite{long_unrealtext_2020}, \href{https://github.com/Belval/TextRecognitionDataGenerator}{TextRecognitionDataGenerator}~\cite{BelvalTe73:online}, \href{https://github.com/BboyHanat/TextGenerator}{Text Generator}~\cite{BboyHana49:online}, \href{https://github.com/wang-tf/Chinese_OCR_synthetic_data}{Chinese OCR synthetic data}~\cite{wangtfCh94:online}, \href{https://github.com/Sanster/text_renderer}{Text Renderer}~\cite{Sanstert53:online} and \href{https://github.com/PaddlePaddle/PaddleOCR/tree/dygraph/StyleText}{the Style-Text package of PaddleOCR}~\cite{wu_editing_2019, PaddleOC59:online}.

As a result of these works, training solely on synthetic data has become a widely accepted practice. Synthetic data alone is sufficient to train state-of-the-art models for the scene text recognition task (tested on real data)~\cite{baek_what_2019, nayef_icdar2019_2019,long_unrealtext_2020}. However, despite good performance on existing real evaluation datasets, some limitations have been identified, including failures on longer characters, smaller sizes and unseen font styles~\cite{chen_text_2020}, and focus on Latin (especially English) or Chinese text~\cite{shi_icdar2017_2018, liu_icdar_2019}.  Recently, more attention has been given to these problems~\cite{islam_survey_2017, chen_text_2020, yan_fast_2018, nayef_icdar2019_2019, shi_script_2016, bustaE2EMLTUnconstrainedEndtoEnd2018}. OmniPrint is helpful to generate small-sized quality text image data, covering extreme distortions in a wide variety of scripts, while giving full control over the choice of distortion parameters, although no special effort has been made, so far, to make such distortions fully realistic to immitate characters in the wild.

\section{The OmniPrint data synthesizer}

\subsection{Overview}

\begin{figure}
    \centering
    \includegraphics[width=\linewidth]{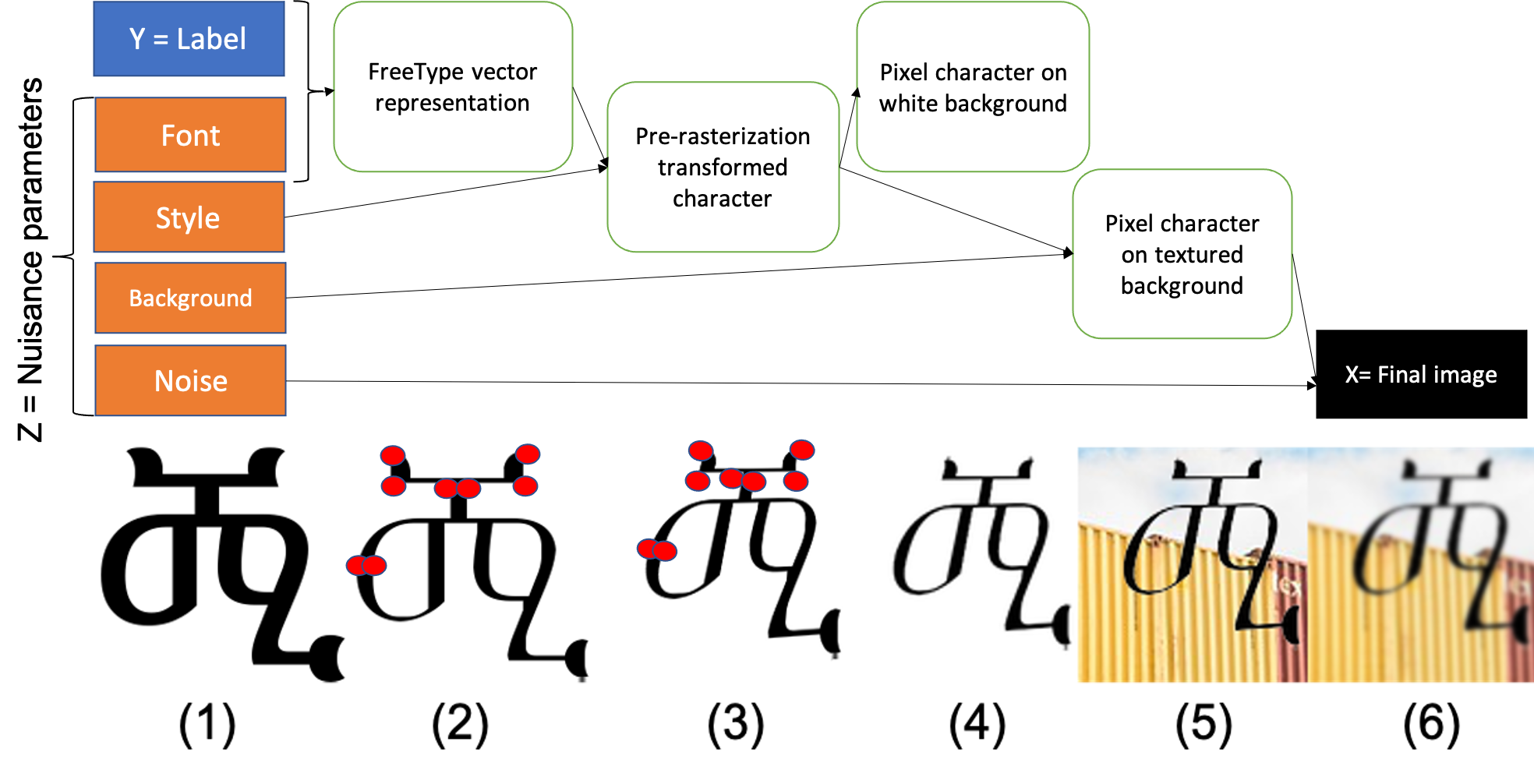}
    \vspace{-8mm}
    \caption{{\bf Basic character image generative process.} The generative process produces images \({\bf X}\) as a function of \({\bf Y}\) (label or character class) and  \({\bf Z}\) (nuisance parameter). Only a subset of anchor point (red dots) are shown in steps (2) and (3). A subset of nuisance parameters are chosen for illustration.}
    \label{fig:architecture}
    \vspace{-3mm}
\end{figure}

OmniPrint is based on the open source software TextRecognitionDataGenerator~\cite{BelvalTe73:online}. While the overall architecture was kept, the software was adapted to meet our required specifications (Table~{\ref{tab:comparisontrdg}} and Figure~\ref{fig:architecture}). To obtain a large number of classes (\({\bf Y}\) labels), we {\bf manually collected and filtered characters} from the Unicode standard in order to form alphabets covering more than 20 languages around the world, these alphabets are further divided into partitions \eg characters from the Oriya script are partitioned into Oriya consonants, Oriya independent vowels and Oriya digits. Nuisance parameters \({\bf Z}\) were decomposed into {\bf Font, Style, Background, and Noise}. To obtain a variety of fonts, we provided an {\bf automatic font collection module}, this module filters problematic fonts and provides fonts' metadata. To obtain a variety of "styles", we substituted the low-level text rendering process by the {\bf FreeType rasterization engine}~\cite{TheFreeT60:online}. This enables {\bf vector-based pre-rasterization transformations}, which are difficult to do with pixel images, such as natural random elastic transformation, stroke width variation and modifications of character proportion (\eg length of ascenders and descenders). We enriched {\bf background generation} with seamless background blending~\cite{10.1145/882262.882269, Gupta16, ankushme88:online}. We proposed a framework for inserting {\bf custom post-rasterization transformations} (\eg perspective transformations, blurring, contrast and brightness variation). Lastly, we implemented {\bf utility} code including dataset formatters, which convert data to AutoML format~\cite{noauthor_automl_nodate} or AutoDL File format~\cite{noauthor_autodl_nodate}, to facilitate the use of such datasets in challenges and benchmarks, and a data loader which generates episodes for meta-learning application.

\begin{table}
  \centering
  \caption{{\bf Comparison of TextRecognitionDataGenerator~\cite{BelvalTe73:online} and OmniPrint.}}
  \label{tab:comparisontrdg}
  \begin{adjustbox}{width=\linewidth}
  \begin{tabular}{lll}
    \toprule
     &  \href{https://github.com/Belval/TextRecognitionDataGenerator}{\underline{TRDG}~\cite{BelvalTe73:online}}  & \underline{OmniPrint} [ours] \\
    \midrule
    Number of characters & 0 & \(12,729\) \\
    Number of words & \(\simeq 11,077,866\) & 0 \\ 
    Number of fonts & 105 & 935 + automatic font collection \\ 
    Pre-rasterization transforms & 0 & 7 (including elastic distortions) \\  
    Post-rasterization transforms & 6 & 15 (+ anti-aliasing rendering) \\
    Foreground & black & color, outline, natural texture \\
    Background & \makecell[l]{speckle noise, quasicrystal, \\white, natural image} & \makecell[l]{same plus seamless blending~\cite{10.1145/882262.882269, Gupta16, ankushme88:online} of \\foreground on background} \\
    Code organization & Transformations hard-coded & Parameter configuration file, Module plug-ins \\
    Dataset formatting & None & \makecell[l]{Metadata recording, standard format support~\cite{noauthor_autodl_nodate, noauthor_automl_nodate}, \\multi-alphabet support, episode generation} \\
    \bottomrule
  \end{tabular}
  \end{adjustbox}
\end{table}



\subsection{Technical aspects of the design}

OmniPrint has been designed to be {\bf extensible}, such that users can easily add new alphabets, new fonts and new transformations into the generation pipeline, see Appendix C, Appendix D and Appendix E. Briefly, here are some highlights of the pipeline of Figure~\ref{fig:architecture}: 

\vspace{-1mm}

\begin{enumerate}
    \item {\bf Parameter configuration file:} We support both TrueType or OpenType font files. Style parameters include rotation angle, shear, stroke width, foreground, text outline and other transformation-specific parameters. 
    \item {\bf FreeType vector representation:} The chosen text, font and style parameters are used as the input to the FreeType rasterization engine~\cite{TheFreeT60:online}. 
    \item {\bf Pre-rasterization transformed character:} FreeType also performs all the pre-rasterization (vector-based) transformations, which include linear transforms, stroke width variation, random elastic transformation and variation of character proportion. The RGB bitmaps output by FreeType are called the foreground layer. 
    \item {\bf Pixel character on white background:} Post-rasterization transformations are applied to the foreground layer. The foreground layer is kept at high resolution at this stage to avoid introducing artifacts. The RGB image is then resized to the desired size with anti-aliasing techniques. The resizing pipeline consists of three steps: (1) applying Gaussian filter to smooth the image; (2) reducing the image by integer times; (3) resizing the image using Lanczos resampling. The second step of the resizing pipeline is an optimization technique proposed by the PIL library~\cite{clark2015pillow}. 
    \item {\bf Pixel character on textured background:} The resized foreground layer is then pasted onto the background at the desired position. 
    \item {\bf Final image:} Some other post-rasterization transformations may be applied after adding the background \eg Gaussian blur of the whole image. Before outputting the synthesized text image, the image mode can be changed if needed (\eg changed to grayscale or binary images).
\end{enumerate}

\vspace{-1mm}

\comment{The above paragraph was changed to an itemized list to facilitate reading.}

Labels \({\bf Y}\) (isolated characters of text) and nuisance parameters \({\bf Z}\) (font, style, background, etc.) are output together with image \({\bf X}\). \({\bf Z}\) serve as "metadata" to help diagnose learning algorithms. The role of \({\bf Y}\) and (a subset of) \({\bf Z}\) may be exchanged to create a variety of classification problems (\eg classifying alphabets or fonts), or regression problems (\eg  predicting rotation angles or shear).

\subsection{Coverage}

We rely on the Unicode 9.0.0 standard~\cite{unicode:online}, which consists of a total of 128172 characters from more than 135 scripts, to identify characters by "code point". A code point is an integer, which represents a single character or part of a character; some code points can be chained to represent a single character \eg the small Latin letter o with circumflex ô can be either represented by a single code point 244 or a sequence of code points (111, 770), where 111 corresponds to the small Latin letter o, 770 means combining circumflex accent. In this work, we use NFC normalized code points~\cite{UAX15Uni97:online} to ensure that each character is uniquely identified. 

We have included 27 scripts: Arabic, Armenian, Balinese, Bengali, Chinese, Devanagari, Ethiopic, Georgian, Greek, Gujarati, Hebrew, Hiragana, Katakana, Khmer, Korean, Lao, Latin, Mongolian, Myanmar, N'Ko, Oriya, Russian, Sinhala, Tamil, Telugu, Thai, Tibetan. For each of these scripts, we manually selected characters. Besides skipping unassigned code points, control characters, incomplete characters, we also filtered Diacritics, tone marks, repetition marks, vocalic modification, subjoined consonants, cantillation marks, etc. For Chinese and Korean characters, we included the most commonly used ones. Details on the selection criteria are given in Appendix F.



The fonts in the first release of OmniPrint have been collected from the Internet. In total, we have selected 12729 characters from 27 scripts (and some special symbols) and 935 fonts. The details of alphabets, fonts can be found in Appendix~C and Appendix~F.

\subsection{Transformations}

\begin{figure}
    \centering
    \includegraphics[width=0.49\linewidth]{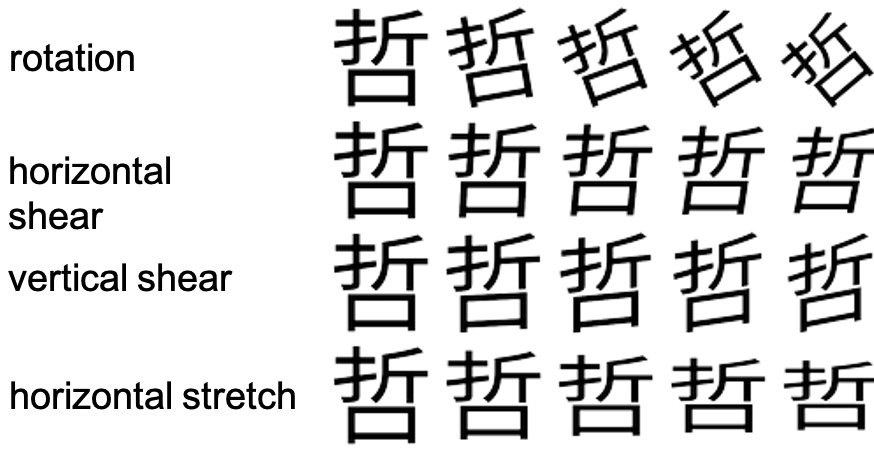} 
    \hspace{1mm}
    \includegraphics[width=0.49\linewidth]{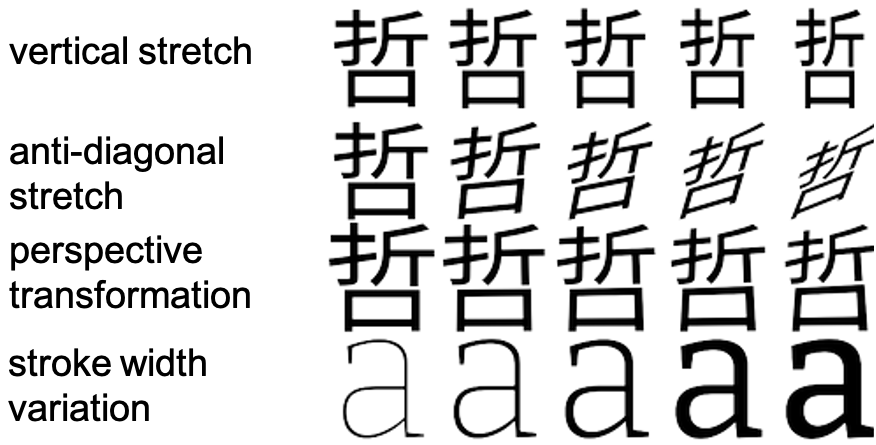}
    \caption{{\bf Some implemented transformations.}}
    \label{fig:transformations}
    \vspace{-3mm}
\end{figure}

Examples of transformations that we implemented are shown in Figure~\ref{fig:transformations}. We are interested in all "label-preserving" transformations on text images as well as their compositions. A transformation is said to be label-preserving if applying it does not alter the semantic meaning of the text image, as interpreted by a human reader. The pre- and post-rasterization transformations that we implemented are detailed in Appendix~D and Appendix~E.

They are classified as {\bf geometric transformations} (number 1-4: each class is a subset of the next class), {\bf local transformations}, and {\bf noises}:


\vspace{-1mm}

\begin{enumerate}
    \item {\bf Isometries: rotation, translation.} Isometries are bijective maps between two metric spaces that preserve distances, they preserve lengths, angles and areas. In our case, rotation has to be constrained to a certain range in order to be label-preserving, the exact range of rotation may vary in function of scripts. Reflection is not desired because it is usually not label-preserving for text images. For human readers, a reflected character may not be acceptable or may even be recognized as another character.
    \item {\bf Similarities: uniform scaling.} Similarities preserve angles and ratios between distances. Uniform scaling includes enlarging or reducing. 
    \item {\bf Affine transformations: shear, stretch.} Affine transformations preserve parallelism. Shear (also known as skew, slant, oblique) can be done either along horizontal axis or vertical axis. Stretch is usually done along the four axes: horizontal axis, vertical axis, main diagonal and anti-diagonal axis. Stretch can be seen as non-uniform scaling. Stretch along horizontal or vertical axis is also referred to as parallel hyperbolic transformation, stretch along main diagonal or anti-diagonal axis is also referred to as diagonal hyperbolic transformation~\cite{simard_transformation_1998}.
    \item {\bf Perspective transformations.} Perspective transformations (also known as homographies or projective transformations) preserve collinearity. This transformation can be used to imitate camera viewpoint \ie 2D projection of 3D world. 
    \item {\bf Local transformations}:   Independent random vibration of the anchor points. Variation of the stroke width \eg thinning or thickening of the strokes.  Variation of character proportion \eg length of ascenders and descenders.
     \item {\bf Noises} related to imaging conditions \eg Gaussian blur, contrast or brightness variation.
\end{enumerate}

\vspace{-1mm}

\section{Use cases}
\subsection{Few-shot learning}
\label{sec:few-shot}

We present a first use case motivated by the organization of the NeurIPS 2021 meta-learning challenge (MetaDL). We use OmniPrint to generate several few-shot learning tasks. Similar datasets are used in the challenge.

\begin{table}
  \centering
  \caption{{\bf Comparison of Omniglot and OmniPrint.}}
  \label{tab:comparisonomniglotOmniPrint}
  \begin{adjustbox}{width=\linewidth}
  \begin{tabular}{lll}
    \toprule
     & \href{https://github.com/brendenlake/omniglot}{\underline{Omniglot}} \cite{lakeHumanlevelConceptLearning2015}    & \href{https://github.com/SunHaozhe/OmniPrint}{\underline{OmniPrint}} [ours] \\
    \midrule
    Total number of unique characters (classes) & 1623 & Unlimited (1409 in our example) \\   
    Total number of examples & 1623\(\times\)20 & Unlimited (1409\(\times\)20 in our example) \\   
    Number of possible alphabets (super-classes)  & 50     & Unlimited (54 in our example) \\
    Scalability of characters and super-classes & No & Yes \\
    Diverse transformations & No & Yes \\    
    Natural background & No & Yes (OmniPrint-meta5 in our exemple) \\
    Possibility of increasing image resolution & No & Yes \\   
    Performance of Prototypical Networks  5-way-1-shot~\cite{snellPrototypicalNetworksFewshot2017} & 98.8\%  & 61.5\%-97.6\%  \\ 
    Performance of MAML 5-way-1-shot~\cite{finnModelAgnosticMetaLearningFast2017} &  98.7\% &  63.4\%-95.0\% \\   
    \bottomrule
  \end{tabular}
  \end{adjustbox}
\end{table}

\begin{table}[]
    \centering
    \caption{{\bf OmniPrint-meta[1-5] datasets} of progressive difficulty. Elastic means random elastic transformations. Fonts are sampled from all the fonts available for each character set. Transformations include random rotation (within -30 and 30 degrees), horizontal shear and perspective transformation. }
    \label{tab:datasetsl}
    \begin{tabular}{cccccc}
    \toprule
    X & Elastic & \# Fonts & Transformations & Foreground & Background \\
    \midrule
    
    1 & \cellcolor[HTML]{C0C0C0}\textbf{Yes} & 1   & No & Black & White \\
    
    2 & \cellcolor[HTML]{C0C0C0}\textbf{Yes} &  \cellcolor[HTML]{C0C0C0}\textbf{Sampled} & No & Black & White \\
    
    3 & \cellcolor[HTML]{C0C0C0}\textbf{Yes} &\cellcolor[HTML]{C0C0C0}\textbf{Sampled} & \cellcolor[HTML]{C0C0C0}\textbf{Yes} & Black & White \\
    
    4 & \cellcolor[HTML]{C0C0C0}\textbf{Yes} & \cellcolor[HTML]{C0C0C0}\textbf{Sampled} & \cellcolor[HTML]{C0C0C0}\textbf{Yes}& \cellcolor[HTML]{C0C0C0}\textbf{Colored} & Colored \\
    
    5 & \cellcolor[HTML]{C0C0C0}\textbf{Yes} & \cellcolor[HTML]{C0C0C0}\textbf{Sampled} & \cellcolor[HTML]{C0C0C0}\textbf{Yes} & \cellcolor[HTML]{C0C0C0}\textbf{Colored} & \cellcolor[HTML]{C0C0C0}\textbf{Textured} \\
    \bottomrule
    \end{tabular}
    \vspace{-3mm}
\end{table}

\begin{figure}[ht]
    \vspace{-0.3cm}
    \centering
    \includegraphics[width=0.8\linewidth]{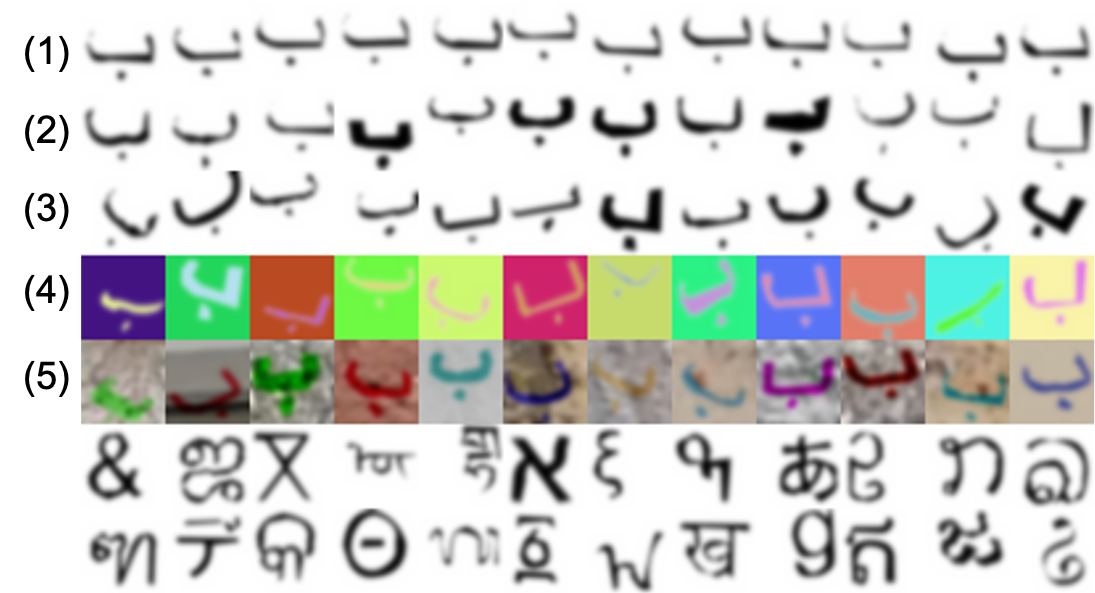}
    \caption{{\bf OmniPrint-meta[1-5] sample data:} Top: The same character of increasing difficulty. Bottom: Examples of characters showing the diversity of the 54 super-classes. }
    \label{fig:omniprintmeta15}
\end{figure} 


Few-shot learning is a ML problem in which new classification problems must be learned "quickly", from just a few training examples per class (shots). This problem is particularly important in domains in which few labeled training examples are available, and/or in which training on new classes must be done quickly episodically (for example if an agent is constantly exposed to new environments). We chose OmniPrint as one application domain of interest for few-shot learning. Indeed, alphabets from many countries are seldom studied and have no dedicated OCR products available. A few-shot learning recognizer could remedy this situation by allowing users to add new alphabets with \eg a single example of each character of a given font, yet generalize to other fonts or styles.

Recently, interest in few-shot learning has been revived (\eg ~\cite{finnModelAgnosticMetaLearningFast2017, snellPrototypicalNetworksFewshot2017, hospedalesMetaLearningNeuralNetworks2020}) and a novel setting proposed. The overall problem is divided into many sub-problems, called episodes. Data are split for each episode into a pair \(\{support~set,~query~set\}\). The support set plays the role of a training set and the query set that of a test set. In the simplified research setting, each episode is supposed to have the same number \(N\) of classes (characters), also called {\bf "ways"}. For each episode, learning machines receive \(K\) training examples per class, also called {\bf "shots"}, in the "support set"; and a number of test examples from the same classes in the "query set". This yields a {\bf \(N\)-way-\(K\)-shot problem}. To perform meta-learning,  data are divided between a {\bf meta-train set} and a {\bf meta-test set}. In the meta-train set, the support and query set labels are visible to learning machines; in contrast, in the meta-test set, only support set labels are visible to learning machines; query set labels are concealed and only used to evaluate performance. In some few-shot learning datasets, classes have hierarchical structures~\cite{triantafillouMetaDatasetDatasetDatasets2020, lakeHumanlevelConceptLearning2015, 2014arXiv1409.0575R} \ie classes sharing certain semantics are grouped into super-classes (which can be thought of as alphabets or partitions of alphabets in the case of OmniPrint). In such cases, episodes can coincide with super-classes, and may have a variable number of "ways".




Using OmniPrint to benchmark few-shot learning methods was inspired by Omniglot~\cite{lakeHumanlevelConceptLearning2015}, a popular benchmark in this field. A typical way of using Omniglot 
is to pool all characters from different alphabets and sample subsets of \(N\) characters to create episodes (\eg \(N=5\) and \(K=1\) results in a 5-way-1-shot problem). While Omniglot has fostered progress, it can hardly push further the state-of-the-art since recent methods, \eg MAML~\cite{finnModelAgnosticMetaLearningFast2017} and Prototypical Networks~\cite{snellPrototypicalNetworksFewshot2017} achieve a classification accuracy of \(98.7\%\) and \(98.8\%\) respectively in the 5-way-1-shot setting. Furthermore, Omniglot was not intended to be a realistic dataset: the characters were drawn online and do not look natural. In contrast OmniPrint provides realistic data with a variability encountered in the real world, allowing us to {\bf create more challenging tasks}.
We compare Omniglot and OmniPrint for few-shot learning benchmarking in Table~\ref{tab:comparisonomniglotOmniPrint}.

As a proof of concept, we created 5 datasets called OmniPrint-meta[1-5] of progressive difficulty, from which few-shot learning tasks can be carved (Table~\ref{tab:datasetsl} and Figure~\ref{fig:omniprintmeta15}). These 5 datasets imitate the setting of Omniglot, for easier comparison and to facilitate replacing it as a benchmark. The OmniPrint-meta[1-5] datasets share the same set of 1409 characters (classes) from 54 super-classes, with 20 examples each, but they {\bf differ in transformations and styles}. Transformations and distortions are cumulated from dataset to dataset, each one including additional transformations to make characters harder to recognize. We synthesized 32\(\times\)32 RGB images of isolated characters. The datasheet for dataset~\cite{gebruDatasheetsDatasets2020} for the OmniPrint-meta[1-5] datasets is shown in Appendix~A.

\deleted{footnote: The 54 super-classes are {\em Arabic, Armenian lowercase, Armenian uppercase ASCII digits, Balinese consonants, Balinese digits, Balinese independent vowels, basic Latin lowercase, basic Latin uppercase, Bengali consonants, Bengali digits, Bengali independent vowels, common punctuations symbols, Devanagari consonants, Devanagari digits, Devanagari independent vowels, Ethiopic digits, Georgian, Greek, Gujarati consonants, Gujarati digits, Gujarati independent vowels, Hebrew, Hiragana, Katakana, Khmer consonants, Khmer digits, Khmer independent vowels, Lao consonants, Lao digits, Mongolian basic letters, Mongolian digits, Myanmar consonants, Myanmar digits, Myanmar independent vowels, N'ko digits, N'ko letters, Oriya consonants, Oriya digits, Oriya independent vowels, Russian, Sinhala astrological digits, Sinhala consonants, Sinhala independent vowels, Tamil consonants, Tamil digits, Tamil independent vowels, Telugu consonants, Telugu digits, Telugu independent vowels, Thai consonants, Thai digits, Tibetan consonants, Tibetan digits.}}







\begin{figure}
    \centering
    \includegraphics[width=\linewidth]{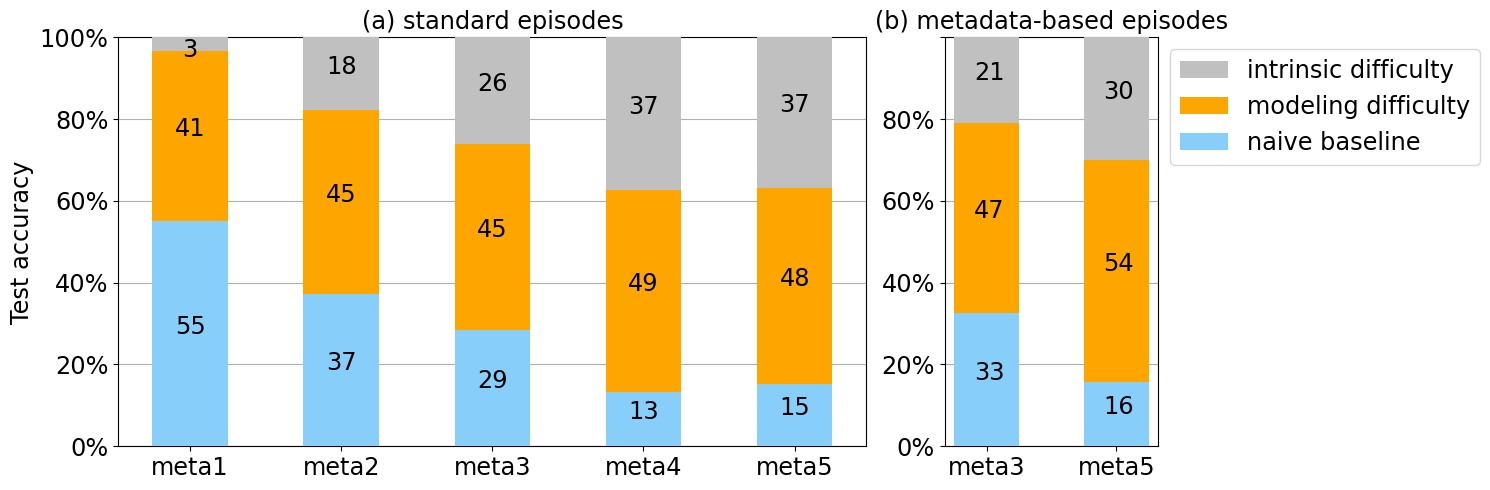}
    \caption{{\bf Difficulty of OmniPrint-meta[1-5] (few-shot learning)}: We averaged the results of \(N\)-way-\(K\)-shot experiments of Table \ref{tab:fewshotresults}. The height of the blue bar represents the performance of the naive baseline (low-end method). The top of the orange bar is the max of the performance of Prototypical Networks and MAML (high-end methods). (a) "Standard" episodes with uniformly sampled rotation and shear. Difficulty progresses from meta1 to meta4, but is (surprisingly) similar between meta4 and meta5. (b) "Metadata" episodes: images within episode share similar rotation \& shear; resulting tasks are easier than corresponding tasks using standard episodes.}
    \label{fig:difficulty}
\end{figure}

We performed a few learning experiments with classical few-shot-learning baseline methods: Prototypical Networks~\cite{snellPrototypicalNetworksFewshot2017} and MAML~\cite{finnModelAgnosticMetaLearningFast2017} (Table~\ref{tab:fewshotresults}). The naive baseline trains a neural network from scratch for each meta-test episode with 20 gradient steps. We split the data into 900 characters for meta-train, 149 characters for meta-validation, 360 characters for meta-test. The model having the highest accuracy on meta-validation episodes during training is selected to be tested on meta-test episodes. Performance is evaluated with the average classification accuracy over 1000 randomly generated meta-test episodes. The reported accuracy and \(95\%\) confidence intervals are computed with 5 independent runs (5 random seeds). The backbone neural network architecture is the same for each combination of method and dataset except for the last fully-connected layer, if applicable. It is the concatenation of three modules of Convolution-BatchNorm-Relu-Maxpool. 
Our findings include that, for 5-way classification of OmniPrint-meta[1-5], MAML outperforms Prototypical Networks, except for OmniPrint-meta1;
for 20-way classification, Prototypical Networks outperforms MAML in easier datasets and are surpassed by MAML for more difficult datasets. One counter-intuitive discovery is that the modeling difficulty estimated from learning machine  performance (Figure ~\ref{fig:difficulty} (a)) does not coincide with human judgement. One would expect that OmniPrint-meta5 should be more difficult than OmniPrint-meta4, because it involves natural backgrounds, making characters visually harder to recognize, but the learning machine results are similar.



\begin{table}
  \caption{{\bf \(N\)-way-\(K\)-shot classification results} on the five OmniPrint-meta[1-5] datasets. }
  \label{tab:fewshotresults}
  \centering
  {\footnotesize 
  \begin{adjustbox}{width=\linewidth}
  \begin{tabular}{lllllll}
    \toprule
    Setting  &   & meta1  & meta2 & meta3 & meta4 & meta5 \\
    \midrule
    \multirow{3}{*}{\shortstack[l]{\(N\)=5\\\(K\)=1}}& 
    Naive & \(66.1\pm 0.7\) & \(43.9 \pm 0.2\) & \(34.9 \pm 0.3\) & \(20.7 \pm 0.1\) & \(22.1 \pm 0.2\) \\
    & Proto~\cite{snellPrototypicalNetworksFewshot2017} &  \(\mathbf{97.6\pm 0.2}\) & \(83.4\pm 0.7\) & \(75.2\pm 1.3\) & \(62.7\pm 0.4\) & \(61.5\pm 0.7\) \\
    & MAML~\cite{finnModelAgnosticMetaLearningFast2017} & \(95.0\pm 0.4\) & \(\mathbf{84.7\pm 0.7}\) & \(\mathbf{76.7\pm 0.4}\) &\(\mathbf{63.4 \pm 1.0}\) & \(\mathbf{63.5\pm 0.8}\) \\
    
    \cmidrule{2-7}
    
    \multirow{3}{*}{\shortstack[l]{\(N\)=5\\\(K\)=5}}& 
    Naive & \(88.7 \pm 0.3\) & \(67.5 \pm 0.5\) & \(52.9 \pm 0.4 \) & \(21.9 \pm 0.1\) & \(26.2 \pm 0.3\) \\
    & Proto~\cite{snellPrototypicalNetworksFewshot2017} & \(\mathbf{99.2\pm 0.1}\) & \(93.6\pm 0.9\) & \(88.6 \pm 1.1\) & \(79.2\pm 1.3\) & \(77.1\pm 1.5\) \\
    & MAML~\cite{finnModelAgnosticMetaLearningFast2017} & \(97.7 \pm 0.2\) & \(\mathbf{93.9\pm 0.5}\) & \(\mathbf{90.4\pm 0.7}\) & \(\mathbf{83.8 \pm 0.5}\) & \(\mathbf{83.8 \pm 0.4}\) \\
    
    \cmidrule{2-7}
    
    \multirow{3}{*}{\shortstack[l]{\(N\)=20\\\(K\)=1}}& 
    Naive & \(25.2 \pm 0.2\) & \(14.3 \pm 0.1\) & \(10.3 \pm 0.1\) & \(5.2 \pm 0.1\) & \(5.8 \pm 0.0\) \\
    & Proto~\cite{snellPrototypicalNetworksFewshot2017} & \(\mathbf{92.2 \pm 0.4}\) & \(\mathbf{66.0 \pm 1.8}\) & \(\mathbf{52.8 \pm 0.7}\) & \(35.6 \pm 0.9\) & \(35.2 \pm 0.7\)  \\
    & MAML~\cite{finnModelAgnosticMetaLearningFast2017} & \(83.3 \pm 0.7\) & \(65.8 \pm 1.3\) & \(52.7 \pm 3.2\) & \(\mathbf{42.0 \pm 0.3}\) & \(\mathbf{42.1 \pm 0.5}\) \\
    
    \cmidrule{2-7}
    
    \multirow{3}{*}{\shortstack[l]{\(N\)=20\\\(K\)=5}}& 
    Naive & \(40.6 \pm 0.1\) & \(23.7 \pm 0.1\) & \(16.0 \pm 0.1\) & \(5.5 \pm 0.0\) & \(6.8 \pm 0.1\) \\
    & Proto~\cite{snellPrototypicalNetworksFewshot2017} & \(\mathbf{97.2 \pm 0.2}\) & \(\mathbf{84.0 \pm 1.1}\) & \(74.1 \pm 0.9\) & \(56.9 \pm 0.4\) & \(54.6 \pm 1.3\)  \\
    & MAML~\cite{finnModelAgnosticMetaLearningFast2017} & \(93.1 \pm 0.3\) & \(83.0 \pm 1.0\) & \(\mathbf{75.9 \pm 1.3}\) & \(\mathbf{61.4 \pm 0.4}\) & \(\mathbf{63.6 \pm 0.5}\)  \\
  \bottomrule
  \end{tabular}
  \end{adjustbox}
  }
  \vspace{-3mm}
\end{table}





\haozhe{Added the following section "Other use cases"}

\haozhe{TODO, update the Abstract to mention these small use cases, or not?}



\subsection{Other meta-learning paradigms}

OmniPrint provides extensively annotated metadata, recording all distortions. Thus more general paradigms of meta-learning (or life-long-learning) can be considered than the few-shot-learning 
setting considered in Section~\ref{sec:few-shot}. Such paradigms may include concept drift or covariate shift. In the former case, distortion parameters, such as rotation or shear, could slowly vary in time; in the latter case episodes could be defined to group examples with similar values of distortion parameters.

To illustrate this idea, we generated episodes differently than in the "standard way"~\cite{snellPrototypicalNetworksFewshot2017, finnModelAgnosticMetaLearningFast2017, vinyalsMatchingNetworksOne2017}. Instead of only varying the subset of classes considered from episode to episode, we also varied  transformation parameters (considered nuisance parameters). This imitates the real-life situation in which data sources and/or recording conditions may vary between data subsets, at either training or test time (or both). We used the two datasets OmniPrint-meta3 and OmniPrint-meta5, described in the previous section, and generated episodes imposing that {\em rotation} and {\em shear} be more similar within episode than between episode (the exact episode generation process and experimental details are provided in Appendix~I). The experimental results, summarized in Figure~\ref{fig:difficulty} (b), show that metadata-based episodes make the problem simpler. The results were somewhat unexpected, but, in retrospect, can be explained by the fact that meta-test tasks are easier to learn, since they are more homogeneous.
This use case of OmniPrint could be developed in various directions, including defining episodes differently at meta-training and meta-test time \eg to study whether algorithms are capable of learning better from more diverse episodes, for fixed meta-test episodes.


\begin{figure}[h]
    \centering
    \begin{minipage}{0.46\linewidth}
      \centering
      \vspace{-1cm}
      \includegraphics[width=\linewidth]{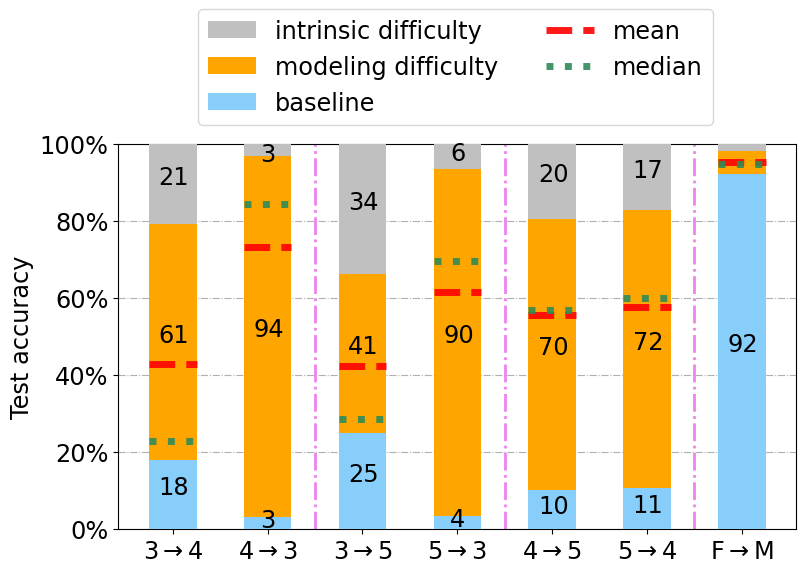}
      \caption{\label{fig:transfer} {\bf Domain adaptation}. \(A\rightarrow B\) means OmniPrint-metaA is source domain and OmniPrint-metaB is target domain, \(A, B \in {3, 4, 5}\). \(\text{F}\rightarrow \text{M}\) means Fake-MNIST is source domain and MNIST is target domain. Mean and median are computed over 5 methods tried. 
     }     
    \end{minipage}%
    \hspace{10mm}
    \begin{minipage}{0.46\linewidth}
      \centering
      \includegraphics[width=\linewidth]{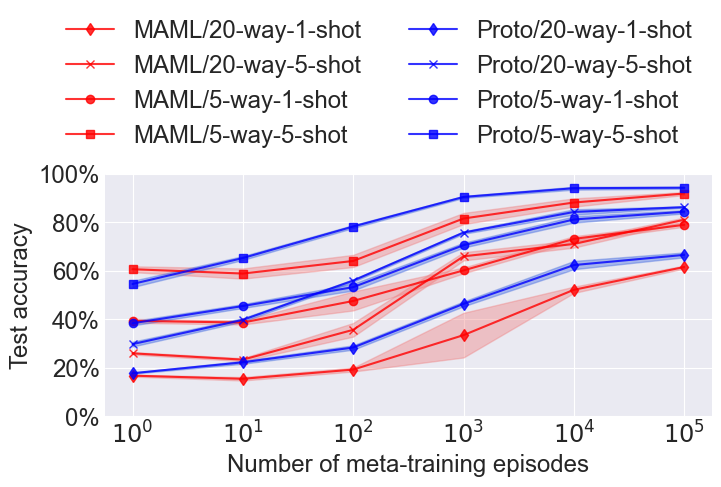}
      \caption{{\bf Influence of the number of meta-training episodes} with a larger version of OmniPrint-meta3. \(95\%\) confidence intervals are computed with 5 random seeds.}
      \label{fig:vary_train_size} 
    \end{minipage}
\end{figure}


\subsection{Influence of the number of meta-training episodes for few-shot learning}


We generated a larger version of OmniPrint-meta3 with 200 images per class (OmniPrint-meta3 has 20 images per class), to study the influence of the number of meta-training episodes. We compared the behavior of  MAML~\cite{finnModelAgnosticMetaLearningFast2017} and Prototypical Network~\cite{snellPrototypicalNetworksFewshot2017}. 
The experiments (Figure~\ref{fig:vary_train_size} and Appendix~J) show that the learning curves cross and Prototypical Network~\cite{snellPrototypicalNetworksFewshot2017} ends with higher performance than MAML~\cite{finnModelAgnosticMetaLearningFast2017} when the number of meta-training episodes increases. Generally Prototypical Network performs better on this larger version of OmniPrint-meta3 than it did on the smaller version. This outlines that changes in experimental settings can reverse conclusions.


\subsection{Domain adaptation}

Since OmniPrint-meta[1-5] datasets share the same label space and only differ in styles and transforms, they lend themselves to benchmarking domain adaptation (DA)~\cite{csurkaDomainAdaptationVisual2017}, one form of transfer learning~\cite{panSurveyTransferLearning2010}. We created a sample DA benchmark, called OmniPrint-metaX-31 based on OmniPrint-meta[3-5] (last 3 datasets). Inspired by Office-31~\cite{hutchisonAdaptingVisualCategory2010}, a popular DA benchmark, we only used 31 randomly sampled characters (out of 1409), and limited ourselves to 3 domains, and 20 examples per class. 
This yields 6 possible DA tasks, for each combinations of domains. We tested each one with the 5 DeepDA unsupervised DA methods~\cite{deepdaJindong}: DAN~\cite{longLearningTransferableFeatures2015, tzengDeepDomainConfusion2014}, DANN~\cite{ganinUnsupervisedDomainAdaptation2015}, DeepCoral~\cite{sunDeepCORALCorrelation2016}, DAAN~\cite{yuTransferLearningDynamic2019} and DSAN~\cite{zhuDeepSubdomainAdaptation2021}. The experimental results are summarized in Figure~\ref{fig:transfer}. More details can be found in Appendix~H. We observe that transfers \(A\rightarrow B\)  when \(A\) is more complex than \(B\) works better than the other way around,
which is consistent with the DA literature~\cite{hoffmanCyCADACycleConsistentAdversarial2017, frenchSelfensemblingVisualDomain2018, liangWeReallyNeed2020}.
The adaptation tasks \(4\rightarrow 5\) and \(5\rightarrow 4\)  are similarly difficult, consistent with Section~\ref{sec:few-shot}. We also observed that when transferring from the more difficult domain to the easier domain, the weakest baseline method (DAN~\cite{longLearningTransferableFeatures2015, tzengDeepDomainConfusion2014}) performs only at chance level, while other methods thrive. We also performed unsupervised DA from a dataset generated with OmniPrint (Fake-MNIST) to MNIST~\cite{lecun2010mnist} (see Appendix~H), The performance of the 5 DeepDA unsupervised DA methods range from 92\% to 98\% accuracy, which is very honorable (current supervised learning results on MNIST are over 99\%).

\subsection{Character image regression tasks}

\haozhe{I did not have time to show the 3 images for Figure 8.}

\begin{figure}[h]
    \centering
    \vspace{-1cm}
    \includegraphics[width=\linewidth]{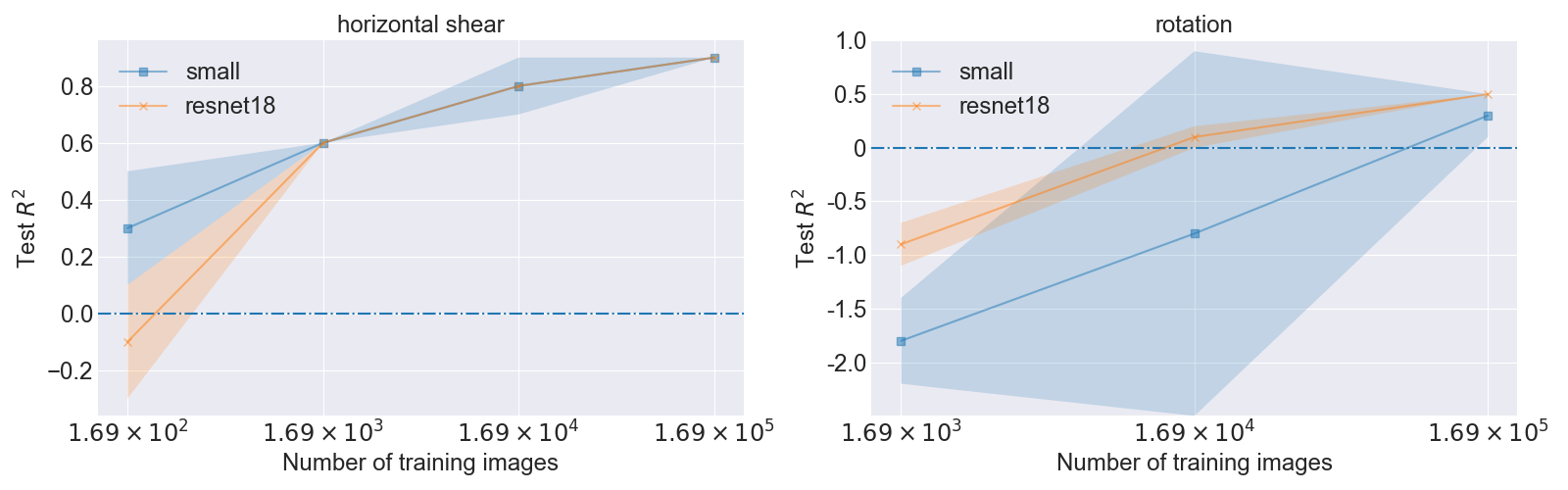}
    \caption{{\bf Regression on text images.} \(95\%\) confidence intervals are computed with 3 random seeds.}
    \label{fig:regression}
\end{figure}

OmniPrint can also be used to generate datasets for regression tasks. We created an example of regression to horizontal shear and rotation. This simulates the problem of detecting variations in style that might have forensic, privacy, and/or fairness implication when characters are handwritten.
OmniPrint could provide training data for bias detection or compensation models. Additionally, shear estimation/detection is one of the preprocessing steps for some OCR methods~\cite{islamSurveyOpticalCharacter2017, beraDatasetNormalizationUnconstrained2019, Harvey647321, pastorImprovingHandwrittenOffline2006, slavikEquivalenceDifferentMethods2001}.

We generated two large datasets which are slightly easier than OmniPrint-meta3. Both datasets contain black-on-white characters (1409 characters with 200 images each). The first dataset has horizontal shear (horizontal shear parameter ranges from -0.8 to 0.8) but not rotation, the second dataset has rotation (rotation ranges from -60 degrees to 60 degrees) but not horizontal shear. Perspective transformations are not used.
We tested two neural networks: A "small" one, concatenating  three modules of Convolution-BatchNorm-Relu-Maxpool, followed by a fully-connected layer with a scalar output (76097 trainable parameters); A "large" one, Resnet18~\cite{heDeepResidualLearning2015} pretrained on ImageNet~\cite{2014arXiv1409.0575R}, of which we trained only the last convolution and fully-connected layers (2360833 trainable parameters). The reported metric is the coefficient of determination \(R^2\). The experiments (Figure~\ref{fig:regression} and Appendix~K) show that horizontal shear is much simpler to predict than rotation.


\haozhe{I do not say that further study should be focused on investigating why rotation are so difficult to predict, because I do not want the reviewers reject the paper for the first round to give us more time......}

\section{Discussion and conclusion}
\label{sec:conclusion}



We developed a new synthetic data generator leveraging existing tools, with a significant number of new features. Datasets generated with OmniPrint retain the simplicity of popular benchmarks such as MNIST or Omniglot. However, while state-of-the-art ML solutions have attained quasi-human performance on MNIST and Omniglot, OmniPrint allows researchers to tune the level of difficulty of tasks, which should foster progress. 

While OmniPrint should provide a useful tool to conduct ML research {\em as is}, it can also be {\em customized} to become an effective OCR research tool. In some respects, OmniPrint goes beyond state-of-the-art software to generate realistic characters. In particular it has the unique capability of incorporating pre-rasterization transformations, allowing users to distort characters by moving anchor points in the original font {\em vector} representation. Still, many synthetic data generators meant to be used for OCR research put emphasis on other aspects, such are more realistic backgrounds, shadows, sensor aberrations, etc., which have not been our priority. Our modular program interface should facilitate such extensions. Another limitation of OmniPrint is that, so far, emphasis has been put on generating isolated characters, although words or sentences can also be generated. Typeset text is beyond the scope of this work.


We do not anticipate any negative societal impact. Much the contrary, OmniPrint should foster research on alphabets that are seldom studied and should allow researchers and developers to expand OCR to many many more languages. Obviously OmniPrint should be responsibly used to balance alphabets from around the world and not discriminate against any culture.

The impact of OmniPrint should go beyond fostering improvement in recognizing isolated printed characters. OmniPrint's data generative process is of the form \({\bf X} = f({\bf Y}, {\bf Z}\)), where \({\bf Y}\) is the class label (character), \({\bf Z}\) encompasses font, style, distortions, background, noises, etc., and \({\bf X}\) is the generated image.  OmniPrint can be used to design tasks in which a label \({\bf Y}\) to be predicted  is entangled with nuisance parameters \({\bf Z}\), resembling other real-world situations in different application domains, to push ML research. This should allow researchers to make progress in a wide variety of problems, whose generative processes are similar (image, video, sound, and text applications, medical diagnoses of genetic disease, analytical chemistry, etc.). Our first meta-learning use cases are a first step in this direction. 

Further work include keeping improving OmniPrint by adding more transformations, and using it in a number of other applications, including image classification benchmarks, data augmentation, study of simulator calibration, bias detection/compensation, modular learning from decomposable/separable problems, recognition of printed characters in the wild and generation of captchas. Our first milestone is using OmniPrint for the NeurIPS2021 meta-learning challenge.











\FloatBarrier


\section*{Acknowledgments and Disclosure of Funding}

\label{sec:acknowledgement}

We gratefully acknowledge many helpful discussions about the project design with our mentors Anne Auger, Feng Han and Romain Egele. We also received useful input from many members of the TAU team of the LISN laboratory, and the MetaDL technical crew: Adrian El Baz, Bin Feng, Jennifer (Yuxuan) He, Jan N. van Rijn, Sebastien Treguer, Ihsan Ullah, Joaquin Vanschoren, Phan Anh Vu, Zhengying Liu, Jun Wan, and Benjia Zhou. OmniPrint is based on the open source software TextRecognitionDataGenerator~\cite{BelvalTe73:online}. We would like to warmly thank all the contributors of this software, especially Edouard Belval. We would like to thank Adrien Pavao for helping providing computing resources. We would also like to thank the reviewers for their constructive suggestions. This work was supported by ChaLearn and the ANR (Agence Nationale de la Recherche, National Agency for Research) under AI chair of excellence HUMANIA, grant number ANR-19-CHIA-0022.











{
\small
\bibliography{main}
}

\end{document}


\maketitle

\appendix

 \section{Datasheet for dataset for OmniPrint-meta[X] datasets}

\newcommand{\dssectionheader}[1]{%
   \noindent\framebox[\columnwidth]{%
      {\textbf{\textcolor{blue}{#1}}}
   }
}

\newcommand{\dsquestion}[1]{%
   {\noindent { \textcolor{blue}{\textbf{#1}}}}
}

\newcommand{\dsquestionex}[2]{%
   {\noindent { \textcolor{blue}{\textbf{#1} #2}}}
}

\newcommand{\dsanswer}[1]{%
   {\noindent {#1} \medskip}
}



\dssectionheader{Motivation}

\dsquestionex{For what purpose was the dataset created?}{Was there a specific task in mind? Was there a specific gap that needed to be filled? Please provide a description.}

\dsanswer{This dataset was created to be a drop-in replacement of Omniglot, which is more challenging. Omniglot can hardly push further the state-of-the-art since recent methods achieved almost perfect performances. Furthermore, Omniglot was not intended to be a realistic dataset: the characters were drawn online and do not look natural. The associated task would be the classical \(N\)-way-\(K\)-shot few-shot classification task~\cite{finnModelAgnosticMetaLearningFast2017, snellPrototypicalNetworksFewshot2017, hospedalesMetaLearningNeuralNetworks2020}.
}


\dsquestion{Who created the dataset (e.g., which team, research group) and on behalf of which entity (e.g., company, institution, organization)?}

\dsanswer{Haozhe Sun created the dataset, under the supervision of Isabelle Guyon. The work was performed at LISN laboratory, Université Paris-Saclay, France, in the TAU team, as part of the HUMANIA project, funded by the French research agency ANR. ChaLearn also supported the development of the software.
}

\dsquestionex{Who funded the creation of the dataset?}{If there is an associated grant, please provide the name of the grantor and the grant name and number.}

\dsanswer{ANR (Agence Nationale de la Recherche, National Agency for Research, \url{https://anr.fr/}), grant number 20HR0134 and ChaLearn (\url{http://www.chalearn.org/}) a 501(c)(3) non-for-profit California organization.
}

\dsquestion{Any other comments?}

\bigskip
\dssectionheader{Composition}

\dsquestionex{What do the instances that comprise the dataset represent (e.g., documents, photos, people, countries)?}{Are there multiple types of instances (e.g., movies, users, and ratings; people and interactions between them; nodes and edges)? Please provide a description.}

\dsanswer{The instances are 32\(\times\)32 RGB images of synthetic printed characters. 
}

\dsquestion{How many instances are there in total (of each type, if appropriate)?}

\dsanswer{OmniPrint-meta[X] is a collection of five datasets. These 5 datasets, called OmniPrint-meta[1-5], share the same set of characters and data split and only differ in transformations and styles. For each OmniPrint-meta[X] dataset, there are 1409 classes (characters) in total. Each class has 20 image instances. In consequence, each OmniPrint-meta[X] dataset has \(1409\times 20=28180\) images. There are \(28180\times 5=140900\) images in total.}

\dsquestionex{Does the dataset contain all possible instances or is it a sample (not necessarily random) of instances from a larger set?}{If the dataset is a sample, then what is the larger set? Is the sample representative of the larger set (e.g., geographic coverage)? If so, please describe how this representativeness was validated/verified. If it is not representative of the larger set, please describe why not (e.g., to cover a more diverse range of instances, because instances were withheld or unavailable).}

\dsanswer{These datasets are synthesized from the data synthesizer OmniPrint, thus they can be viewed as a sample of instances from all the possible images given the nuisance parameters (fonts, styles, noises, etc.). OmniPrint-meta[X] are representative of such images because the synthesis parameters of each instance were uniformly sampled, no further selection was performed. The involved scripts are Arabic, Armenian, Balinese, Latin, Bengali, Devanagari, Ethiopic, Georgian, Greek, Gujarati, Hebrew, Hiragana, Katakana, Khmer, Lao, Mongolian, Myanmar, N'Ko, Oriya, Russian, Sinhala, Tamil, Telugu, Thai and Tibetan.
}




\dsquestionex{What data does each instance consist of?}{"Raw" data (e.g., unprocessed text or images) or features? In either case, please provide a description.}

\dsanswer{Each instance is a 32\(\times\)32 RGB image. Each image contains one single character from a certain script, rendered in a particular way (background, foreground, distortions, noises).
}


\dsquestionex{Is there a label or target associated with each instance?}{If so, please provide a description.}

\dsanswer{Yes, there is a label (character) associated with each instance. Furthermore, the metadata is provided for each instance, which can also serve as labels for specific tasks. The metadata includes \eg the font, background, stroke width (if applicable), blur radius, margins, rotation angle, shear, text color, etc., and the alphabet of the character.
}


\dsquestionex{Is any information missing from individual instances?}{If so, please provide a description, explaining why this information is missing (e.g., because it was unavailable). This does not include intentionally removed information, but might include, e.g., redacted text.}

\dsanswer{No. All of the metadata is provided for each instance.
}

\dsquestionex{Are relationships between individual instances made explicit (e.g., users’ movie ratings, social network links)?}{If so, please describe how these relationships are made explicit.}

\dsanswer{All relationships are contained in the labels and metadata, all provided.
}





\dsquestionex{Are there recommended data splits (e.g., training, development/validation, testing)?}{If so, please provide a description of these splits, explaining the rationale behind them.}

\dsanswer{Yes, there is a recommended data split in the context of \(N\)-way-\(K\)-shot learning, between meta-train, meta-validation and meta-test. For each of the 5 OmniPrint-meta[X] datasets, there are 1409 classes (characters), each class contains 20 image instances. The first 900 classes belong to meta-train, then 149 classes belong to meta-validation, the last 360 classes belong to meta-test. This data split is chosen in order to imitate the proportion of meta-train/meta-validation/meta-test of the popular Vinyals split~\cite{vinyalsMatchingNetworksOne2017} of Omniglot~\cite{lakeHumanlevelConceptLearning2015}. The recommended data split is provided via a data loader which forms the episodes of few-shot learning.
}

\dsquestionex{Are there any errors, sources of noise, or redundancies in the dataset?}{If so, please provide a description.}

\dsanswer{We intentionally introduced various transformations and noises to each image instance. The transformation parameter space is large so there is little chance that two instances are identical.
}



\dsquestionex{Is the dataset self-contained, or does it link to or otherwise rely on external resources (e.g., websites, tweets, other datasets)?}{If it links to or relies on external resources, a) are there guarantees that they will exist, and remain constant, over time; b) are there official archival versions of the complete dataset (i.e., including the external resources as they existed at the time the dataset was created); c) are there any restrictions (e.g., licenses, fees) associated with any of the external resources that might apply to a future user? Please provide descriptions of all external resources and any restrictions associated with them, as well as links or other access points, as appropriate.}

\dsanswer{The 5 datasets OmniPrint-meta[X] are self-contained. They will exist, and remain constant, over time once we release them after the NeurIPS 2021 meta-learning challenge.
}




\dsquestionex{Does the dataset contain data that might be considered confidential (e.g., data that is protected by legal privilege or by doctor-patient confidentiality, data that includes the content of individuals’ non-public communications)?}{If so, please provide a description.}

\dsanswer{The OmniPrint-meta[X] datasets were considered confidential before the NeurIPS 2021 meta-learning challenge, they have been publicly released.
}


\dsquestionex{Does the dataset contain data that, if viewed directly, might be offensive, insulting, threatening, or might otherwise cause anxiety?}{If so, please describe why.}

\dsanswer{No.
}


\dsquestionex{Does the dataset relate to people?}{If not, you may skip the remaining questions in this section.}

\dsanswer{No.
}

\dsquestionex{Does the dataset identify any subpopulations (e.g., by age, gender)?}{If so, please describe how these subpopulations are identified and provide a description of their respective distributions within the dataset.}

\dsanswer{No.
}

\dsquestionex{Is it possible to identify individuals (i.e., one or more natural persons), either directly or indirectly (i.e., in combination with other data) from the dataset?}{If so, please describe how.}

\dsanswer{No.
}

\dsquestionex{Does the dataset contain data that might be considered sensitive in any way (e.g., data that reveals racial or ethnic origins, sexual orientations, religious beliefs, political opinions or union mem- berships, or locations; financial or health data; biometric or genetic data; forms of government identification, such as social security numbers; criminal history)?}{If so, please provide a description.}

\dsanswer{No.
}

\dsquestion{Any other comments?}



\dsanswer{
}

\bigskip
\dssectionheader{Collection Process}

\dsquestionex{How was the data associated with each instance acquired?}{Was the data directly observable (e.g., raw text, movie ratings), reported by subjects (e.g., survey responses), or indirectly inferred/derived from other data (e.g., part-of-speech tags, model-based guesses for age or language)? If data was reported by subjects or indirectly inferred/derived from other data, was the data validated/verified? If so, please describe how.}

\dsanswer{Each instance is synthesized by OmniPrint. Each instance is an image and is directly observable. 
}

\dsquestionex{What mechanisms or procedures were used to collect the data (e.g., hardware apparatus or sensor, manual human curation, software program, software API)?}{How were these mechanisms or procedures validated?}

\dsanswer{The data are synthesized using the data synthesizer OmniPrint. The involved Unicode characters were manually selected from the Unicode standard, which constitutes a set of characters from several languages around the world. The involved fonts were downloaded from a manually-defined list of URLs, the downloaded fonts were then filtered by a python program in order to filter corrupted fonts. Several distortions and noises were involved, including affine and perspective transformations, random elastic transformations, natural background, foreground text filling, etc.
}



\dsquestion{If the dataset is a sample from a larger set, what was the sampling strategy (e.g., deterministic, probabilistic with specific sampling probabilities)?}

\dsanswer{The data is synthesized by a data synthesizer OmniPrint. The sampling is uniformly random in the given transformation parameter space.
}


\dsquestion{Who was involved in the data collection process (e.g., students, crowdworkers, contractors) and how were they compensated (e.g., how much were crowdworkers paid)?}

\dsanswer{The data is synthesized by a computer software. However the design and implementation of the software, the choice of characters and fonts involve the authors of this paper.
}


\dsquestionex{Over what timeframe was the data collected?}{Does this timeframe match the creation timeframe of the data associated with the instances (e.g., recent crawl of old news articles)? If not, please describe the timeframe in which the data associated with the instances was created.}

\dsanswer{The five datasets were synthesized on May 22, 2021.
}


\dsquestionex{Were any ethical review processes conducted (e.g., by an institutional review board)?}{If so, please provide a description of these review processes, including the outcomes, as well as a link or other access point to any supporting documentation.}

\dsanswer{N/A
}

\dsquestionex{Does the dataset relate to people?}{If not, you may skip the remainder of the questions in this section.}

\dsanswer{No.
}

\dsquestion{Did you collect the data from the individuals in question directly, or obtain it via third parties or other sources (e.g., websites)?}

\dsanswer{N/A
}

\dsquestionex{Were the individuals in question notified about the data collection?}{If so, please describe (or show with screenshots or other information) how notice was provided, and provide a link or other access point to, or otherwise reproduce, the exact language of the notification itself.}

\dsanswer{N/A
}

\dsquestionex{Did the individuals in question consent to the collection and use of their data?}{If so, please describe (or show with screenshots or other information) how consent was requested and provided, and provide a link or other access point to, or otherwise reproduce, the exact language to which the individuals consented.}

\dsanswer{N/A
}

\dsquestionex{If consent was obtained, were the consenting individuals provided with a mechanism to revoke their consent in the future or for certain uses?}{If so, please provide a description, as well as a link or other access point to the mechanism (if appropriate).}

\dsanswer{N/A
}

\dsquestionex{Has an analysis of the potential impact of the dataset and its use on data subjects (e.g., a data protection impact analysis)been conducted?}{If so, please provide a description of this analysis, including the outcomes, as well as a link or other access point to any supporting documentation.}

\dsanswer{N/A
}

\dsquestion{Any other comments?}

\dsanswer{
}

\bigskip
\dssectionheader{Preprocessing/cleaning/labeling}

\dsquestionex{Was any preprocessing/cleaning/labeling of the data done (e.g., discretization or bucketing, tokenization, part-of-speech tagging, SIFT feature extraction, removal of instances, processing of missing values)?}{If so, please provide a description. If not, you may skip the remainder of the questions in this section.}

\dsanswer{No preprocessing/cleaning/labeling was performed. The datasets are made available as they were synthesized. No feature extraction or removal of instances was done.
}

\dsquestionex{Was the “raw” data saved in addition to the preprocessed/cleaned/labeled data (e.g., to support unanticipated future uses)?}{If so, please provide a link or other access point to the “raw” data.}

\dsanswer{N/A
}

\dsquestionex{Is the software used to preprocess/clean/label the instances available?}{If so, please provide a link or other access point.}

\dsanswer{N/A
}

\dsquestion{Any other comments?}

\dsanswer{
}

\bigskip
\dssectionheader{Uses}

\dsquestionex{Has the dataset been used for any tasks already?}{If so, please provide a description.}

\dsanswer{No, however a variant  of these datasets will be used by the NeurIPS 2021 meta-learning challenge. 
}




\dsquestionex{Is there a repository that links to any or all papers or systems that use the dataset?}{If so, please provide a link or other access point.}

\dsanswer{Yes, the link is \href{https://github.com/SunHaozhe/OmniPrint-datasets}{https://github.com/SunHaozhe/OmniPrint-datasets}. This repository is also used to announce any necessary information related to the OmniPrint datasets \eg potential changes of the dataset hosting address.
}



\dsquestion{What (other) tasks could the dataset be used for?}

\dsanswer{Besides few-shot learning classification tasks, the five OmniPrint-meta[X] datasets can be used for classification tasks of a large number of characters, and for transfer learning (each dataset being used either as a source domain or a target domain). Furthermore, as the metadata can serve as labels, other kinds of classification or regression problems can also be considered \eg classification of fonts, classification of languages, regression of rotation angle, regression of horizontal shear, etc. Finally, the datasets can be used to study disentangling the label (class character) from the nuisance variables (font, style, distortions).
}


\dsquestionex{Is there anything about the composition of the dataset or the way it was collected and preprocessed/cleaned/labeled that might impact future uses?}{For example, is there anything that a future user might need to know to avoid uses that could result in unfair treatment of individuals or groups (e.g., stereotyping, quality of service issues) or other undesirable harms (e.g., financial harms, legal risks) If so, please provide a description. Is there anything a future user could do to mitigate these undesirable harms?}

\dsanswer{The datasets can be used without further considerations. 
}



\dsquestionex{Are there tasks for which the dataset should not be used?}{If so, please provide a description.}

\dsanswer{Not that we know of.
}



\dsquestion{Any other comments?}

\bigskip
\dssectionheader{Distribution}

\dsquestionex{Will the dataset be distributed to third parties outside of the entity (e.g., company, institution, organization) on behalf of which the dataset was created?}{If so, please provide a description.}

\dsanswer{The datasets are made available to everyone via the Internet.
}

\dsquestionex{How will the dataset will be distributed (e.g., tarball on website, API, GitHub)?}{Does the dataset have a digital object identifier (DOI)?}

\dsanswer{The OmniPrint-meta[X] datasets are publicly released via \href{https://www.kaggle.com/datasets}{Kaggle Datasets}. The digital object identifier (DOI) is 10.34740/kaggle/dsv/2763401. The access information and any necessary updates are announced via \href{https://github.com/SunHaozhe/OmniPrint-datasets}{https://github.com/SunHaozhe/OmniPrint-datasets}. 
}




\dsquestion{When will the dataset be distributed?}

\dsanswer{The datasets have been released after the NeurIPS 2021 meta-learning challenge. 
}


\dsquestionex{Will the dataset be distributed under a copyright or other intellectual property (IP) license, and/or under applicable terms of use (ToU)?}{If so, please describe this license and/or ToU, and provide a link or other access point to, or otherwise reproduce, any relevant licensing terms or ToU, as well as any fees associated with these restrictions.}

\dsanswer{The datasets OmniPrint-meta[1-5] are distributed via Kaggle datasets. They are licensed under a Creative Commons license CC BY 4.0 \href{https://creativecommons.org/licenses/by/4.0/}{https://creativecommons.org/licenses/by/4.0/}. This comes with the following guarantee disclaimer: Unless otherwise separately undertaken by the Licensor, to the extent possible, the Licensor offers the Licensed Material as-is and as-available, and makes no representations or warranties of any kind concerning the Licensed Material, whether express, implied, statutory, or other. This includes, without limitation, warranties of title, merchantability, fitness for a particular purpose, non-infringement, absence of latent or other defects, accuracy, or the presence or absence of errors, whether or not known or discoverable. Where disclaimers of warranties are not allowed in full or in part, this disclaimer may not apply to You. To the extent possible, in no event will the Licensor be liable to You on any legal theory (including, without limitation, negligence) or otherwise for any direct, special, indirect, incidental, consequential, punitive, exemplary, or other losses, costs, expenses, or damages arising out of this Public License or use of the Licensed Material, even if the Licensor has been advised of the possibility of such losses, costs, expenses, or damages. Where a limitation of liability is not allowed in full or in part, this limitation may not apply to You.
}




\dsquestionex{Have any third parties imposed IP-based or other restrictions on the data associated with the instances?}{If so, please describe these restrictions, and provide a link or other access point to, or otherwise reproduce, any relevant licensing terms, as well as any fees associated with these restrictions.}

\dsanswer{No.
}


\dsquestionex{Do any export controls or other regulatory restrictions apply to the dataset or to individual instances?}{If so, please describe these restrictions, and provide a link or other access point to, or otherwise reproduce, any supporting documentation.}

\dsanswer{No.
}

\dsquestion{Any other comments?}

\dsanswer{
}

\bigskip
\dssectionheader{Maintenance}

\dsquestion{Who is supporting/hosting/maintaining the dataset?}

\dsanswer{The authors of this paper are responsible for supporting the datasets. 
}





\dsquestion{How can the owner/curator/manager of the dataset be contacted (e.g., email address)?}

\dsanswer{The preferred way to contact the maintainers is to raise issues on \href{https://github.com/SunHaozhe/OmniPrint-datasets}{https://github.com/SunHaozhe/OmniPrint-datasets}. In case of emergency, the authors of this paper can be contacted via email: omniprint@chalearn.org.
}



\dsquestionex{Is there an erratum?}{If so, please provide a link or other access point.}

\dsanswer{Any necessary information or updates will be accessible via \href{https://github.com/SunHaozhe/OmniPrint-datasets}{https://github.com/SunHaozhe/OmniPrint-datasets}.
}

\dsquestionex{Will the dataset be updated (e.g., to correct labeling errors, add new instances, delete instances)?}{If so, please describe how often, by whom, and how updates will be communicated to users (e.g., mailing list, GitHub)?}

\dsanswer{No. New needs will be met by synthesizing new datasets.
}

\dsquestionex{If the dataset relates to people, are there applicable limits on the retention of the data associated with the instances (e.g., were individuals in question told that their data would be retained for a fixed period of time and then deleted)?}{If so, please describe these limits and explain how they will be enforced.}

\dsanswer{N/A
}

\dsquestionex{Will older versions of the dataset continue to be supported/hosted/maintained?}{If so, please describe how. If not, please describe how its obsolescence will be communicated to users.}

\dsanswer{Any necessary information or updates will be accessible via \href{https://github.com/SunHaozhe/OmniPrint-datasets}{https://github.com/SunHaozhe/OmniPrint-datasets}.
}

\dsquestionex{If others want to extend/augment/build on/contribute to the dataset, is there a mechanism for them to do so?}{If so, please provide a description. Will these contributions be validated/verified? If so, please describe how. If not, why not? Is there a process for communicating/distributing these contributions to other users? If so, please provide a description.}

\dsanswer{Users are free to extend or augment the dataset for their purposes. They can also use the data synthesizer OmniPrint to directly synthesize new datasets.
}

\dsquestion{Any other comments?}

\dsanswer{
}


\clearpage

\section{Experimental details of the few-shot learning use case}
\label{sec:experimentaldetails}

This section provides the experimental details of Section~4.1 of the main paper.

\subsection{Data split}

We split the data into 900 characters for meta-train, 149 characters for meta-validation, 360 characters for meta-test. The full details are provided with the code. The implementation of the few-shot learning data loader which forms the few-shot learning episodes is inspired by \cite{MAML_Pytorch_Liangqu_Long} which is under MIT License.

\subsection{Evaluation and reproducibility}

MAML~\cite{finnModelAgnosticMetaLearningFast2017} and Prototypical Networks~\cite{snellPrototypicalNetworksFewshot2017} were trained during 300 epochs, where each epoch is defined to be 6 batches of episodes, each batch contains 32 episodes. During meta-training, the model checkpoints were evaluated on meta-validation episodes every 5 epochs. Only the checkpoint that has the highest accuracy on meta-validation episodes during training is selected to be tested on meta-test episodes. 

The backbone neural network architecture is the same for each combination of method and dataset except for the last fully-connected layer, if applicable. It is the concatenation of three modules of Convolution-BatchNorm-Relu-Maxpool. 

The metric of interest is the average classification accuracy of 1000 randomly generated meta-test episodes (of the best checkpoint on meta-validation episodes). The reported accuracy and \(95\%\) confidence intervals in the main paper are computed with 5 independent runs (5 random seeds). The random seeds were fixed in advance, no cherry-picking was performed afterwards.

\subsection{Baseline implementation and compute resources}

The implementation of MAML baseline~\cite{finnModelAgnosticMetaLearningFast2017} uses the Higher library~\cite{grefenstetteGeneralizedInnerLoop2019} of PyTorch~\cite{NEURIPS2019_9015}. It is adapted from \cite{higherma24:online} which is under Apache License Version 2.0. The implementation of Prototypical Networks~\cite{snellPrototypicalNetworksFewshot2017} is adapted from \cite{orobixPr43:online} which is under MIT License. 

The experiments were run on an internal cluster which is managed through SLURM~\cite{Jette02slurm:simple}. The involved GPUs are Tesla K80, Tesla V100-PCIE-32GB, Tesla V100-SXM2-32GB. Each run uses one single GPU. The experiments involve 5 datasets OmniPrint-meta[1-5], 3 baseline methods (MAML, PyTorch, Naive), 4 settings (5-way-1-shot, 5-way-5-shot, 20-way-1-shot, 20-way-5-shot), 5 random seeds. The total amount of computation time is about 280 hours.

\subsection{Hyperparameters}

The baseline methods used the default or recommended hyperparameters of the original paper/code. A small number of hyperparameters \eg learning rates, were adjusted according to preliminary experiments. No large-scale hyperparameter optimization was performed.

While the full details are provided with the code, we highlight some important hyperparameters: 

\begin{itemize}
    \item {\bf MAML}~\cite{finnModelAgnosticMetaLearningFast2017} 5 inner steps were used for meta-train, meta-validation and meta-test. The meta learner is optimized using Adam~\cite{kingmaAdamMethodStochastic2017} with the learning rate \(10^{-3}\). The inner loops were optimized using SGD~\cite{2016arXiv160904747R} with the learning rate \(10^{-1}\).
    \item {\bf Prototypical Networks}~\cite{snellPrototypicalNetworksFewshot2017} By following the original paper, each meta-train episode is a \(60\)-way-\(K\)-shot regardless the meta-validation/meta-test setting. No learning rate decay was used. The backbone neural network was optimized using Adam~\cite{kingmaAdamMethodStochastic2017} with the learning rate \(5\times 10^{-4}\).
    \item {\bf Naive} The neural network for each meta-test episode was trained from scratch (random initialization) with 20 gradient steps. It was optimized using Adam~\cite{kingmaAdamMethodStochastic2017} with the learning rate \(10^{-4}\).
\end{itemize}

\subsection{Data synthesis}

The background images used for OmniPrint-meta5 dataset were taken using a personal mobile phone.

\section{Fonts}


Fonts are usually protected under their own licenses. We do not provide any warranty for this. Please be aware that some fonts cannot be redistributed or modified. This is the reason why we do not redistribute fonts with our code. However, we provide the font preparation scripts that we used. These fonts were downloaded from a manually-collected list of URLs. 

We provide the font preparation scripts. If some URLs fail, please consider re-run the scripts at a later time (possibly related to network problems). If some URLs continue to fail, please contact the authors of this paper (via GitHub Issues page or via email: omniprint@chalearn.org). On the other hand, the users are free to collect their own set of fonts depending on their needs.

We gathered a list of URLs and prepared scripts which automatically download, filter and format the fonts. These scripts also record metadata of these fonts. The workflow of the font preparation scripts can be summarized into 2 stages: 

\begin{itemize}
    \item {\bf Downloading} Download files from the given URLs, logs will be generated to keep track of potential failures. After unzipping, reformat file names which handles decoding error, converts file names to lower case, remove invalid symbols and translate Chinese file names. Generate metadata about sources of each font: some URLs contain several fonts, the same font can also be downloaded from different URLs. 
    \item {\bf Building} Filter out corrupted or unwanted fonts and move all font files to the dedicated directory. Move all license files to the dedicated directory. Build the so-called index files for each alphabet. Each alphabet has an index file which contains a list of fonts that support all of the characters it contains. Generate the lists of variable fonts and save the metadata of fonts \eg family name, style name, the range of possible stroke width (if any), etc. into a csv file.
\end{itemize}

Importing new fonts is easy in OmniPrint. 

\begin{enumerate}
    \item Move new fonts to the directory {\it fonts/fonts/}
    \item Optionally, update the index file under the directory {\it fonts/index/} if users want to randomly select fonts 
    \item Optionally, update the metadata of fonts under the directory {\it fonts/metadata/}
    \item Users should not forget to include license files in the directory {\it fonts/licenses/}
\end{enumerate}

If users want to collect their own set of fonts, please be aware that some fonts can produce false rendering (empty image, square as a placeholder or even random symbols) without reporting any warnings or errors. 


\section{Pre-rasterization transformations}

\begin{figure}[htb]
    \centering
    \includegraphics[width=\linewidth]{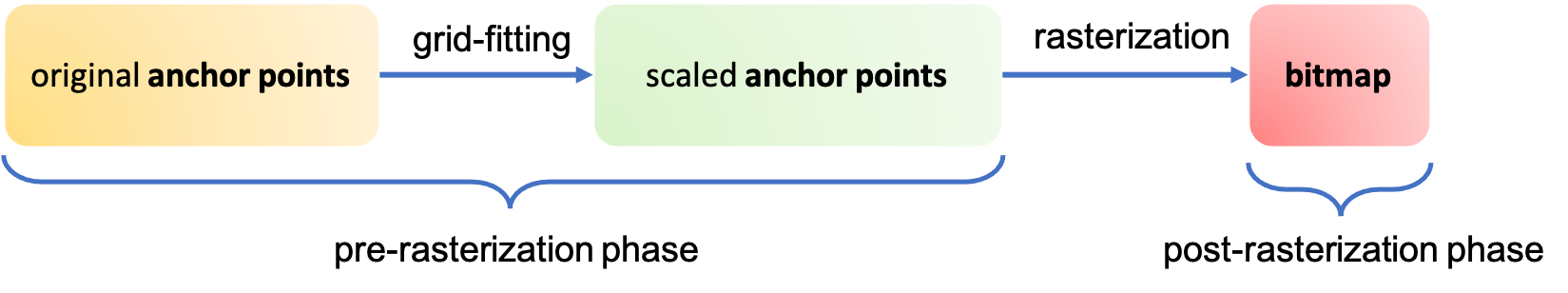}
    \caption{{\bf Conversion process from TrueType/OpenType fonts to digital images.} In OmniPrint, pre-rasterization elastic transformation is performed on the original anchor points (yellow), linear transformations of anchor points are performed on the scaled anchor points (green).}
    \label{fig:font_lifecycle}
\end{figure}

The rendering process of modern digital fonts (TrueType/OpenType) is divided into two phases by the rasterization. Digital fonts are originally stored as anchor points expressed in font units within the EM square. Before being able to be rendered into bitmaps, the anchor points are scaled to be aligned with the device pixel grid. The grid-fitting (also called hinting) and rasterization are performed by the FreeType engine (Figure~\ref{fig:font_lifecycle}). 

Pre-rasterization transformations refer to direct manipulation of the anchor points of the digital font files. Modern fonts (\eg TrueType or OpenType) are made of straight line segments and quadratic Bézier curves, connecting anchor points. OmniPrint uses the low-level Freetype font rasterization engine~\cite{TheFreeT60:online} (Python binding~\cite{rougierf53:online} which is under BSD license), which makes direct manipulation of anchor points possible. With pre-rasterization transformations, one can deform the characters without incurring aberrations due to aliasing and generate some local deformations that would be difficult to achieve with post-rasterization transformations (digital image processing) \ie natural elastic transformation, variation of character proportion, structured deformation of specific characters, etc. 

\begin{algorithm}
\caption{Pre-rasterization elastic transformation}
\label{alg:preelastic}
\SetAlgoLined
\DontPrintSemicolon
\KwIn{A sequence of characters \(S\), a digital font \(F\), a probability distribution \(D\)}
\KwOut{Rendered text image \(I\)}
\tcp*[l]{\(C\) denotes characters, \(P\) denotes anchor points, the function {\it load} loads the {\bf initial} anchor points of a digital font for a certain character. The function {\it enumerate} returns the index as well as the value of an array.}
\tcp*[l]{First pass to compute bounding box of the sequence}
xmin, xmax, ymin, ymax = 0, 0, 0, 0 \;
Initialize cache \tcp*[l]{In order to save random vibration} 
\For{C in S}{
    \For{P in load(C, F)}{
    	xdelta \(\sim D\) \;
    	ydelta \(\sim D\) \;
    	\(P\).x \(\leftarrow\) \(P\).x + xdelta \;
    	\(P\).y \(\leftarrow\) \(P\).y + ydelta \;
    	cache.append(\,(xdelta, ydelta)\,) \;
    	xmin, xmax, ymin, ymax \(\leftarrow\) update(xmin, xmax, ymin, ymax, \(P\)) 
    }
}
\(I \leftarrow\) build\_image(xmin, xmax, ymin, ymax)\;
\tcp*[l]{Second pass to render text}
\For{i, C in enumerate(S)}{
    \For{j, P in enumerate(\,load(C, F)\,)}{
    	\(P\).x \(\leftarrow\) \(P\).x + cache[\(i\)][\(j\)][0] \;
    	\(P\).y \(\leftarrow\) \(P\).y + cache[\(i\)][\(j\)][1] \;
    }
    \(I \leftarrow\) fill\_image(\(I, C\))
}
\end{algorithm}

The implemented pre-rasterization transformations are listed as follows:

\begin{itemize}
    \item {\bf Elastic transformation (pre-rasterization)} corresponds to random vibration of independent anchor points. The pseudocode is shown in Algorithm~\ref{alg:preelastic}. Of note is that elastic transformations are implemented in both pre-rasterization phase and post-rasterization phase, which can also be used together. All the elastic transformations mentioned in the main paper refer to pre-rasterization elastic transformation.
    \item {\bf Stroke width variation} Variation of the stroke width \eg thinning or thickening of the strokes. Only variable fonts support stroke width variation, each variable font has its own continuous range of permissible stroke width.
    \item {\bf Variation of character proportion} \eg variation of length of ascenders and descenders by some font units.
    \item {\bf Linear transformations} Rotation, shear, scaling, stretch are assembled into a \(2\times 2\) matrix, see Equation~\ref{eq:linear_transformation}. \(\theta\) denotes the angle (in degree) of counter clockwise rotation, \(\lambda_1, \lambda_2\) denote the shear parameters along horizontal axis and vertical axis respectively, \(s_1, s_2\) denote the scaling (stretch) parameters along horizontal axis and vertical axis respectively. If \(s_1=s_2\), this corresponds to a scaling operation, otherwise this corresponds to a stretch operation along horizontal or vertical axes. The stretch along main diagonal axis and anti-diagonal axis by setting \(\beta = \gamma \in \mathbb{R}\) or \(\lambda_1 = \lambda_2 \in \mathbb{R}\)~\cite{simard_transformation_1998}. The four parameters \(\alpha, \beta, \gamma, \delta\) allow inserting an arbitrary linear transform into the default linear transformation pipeline. Users are also allowed to directly set the values of \(a, b, d, e\) \ie the composed linear transformation matrix \(L\). 
\end{itemize}


\vspace{-0.5cm}

\begin{equation}
    \label{eq:linear_transformation}
    \begin{split}
        L 
        &= 
        \begin{pmatrix}
            a & b \\
            d & e
        \end{pmatrix} \\
        &= 
        \begin{pmatrix}
            \cos{\theta} & -\sin{\theta} \\
            \sin{\theta} & \cos{\theta} 
        \end{pmatrix}
        \begin{pmatrix}
            1 & \lambda_1 \\
            \lambda_2 & 1
        \end{pmatrix}
        \begin{pmatrix}
            \alpha & \beta \\
            \gamma & \delta 
        \end{pmatrix}
        \begin{pmatrix}
            s_1 & 0 \\
            0 & s_2
        \end{pmatrix}  \\
        &= 
        \begin{pmatrix}
            s_1 ((\alpha + \gamma \lambda_1) \cos{\theta} - (\alpha \lambda_2 + \gamma) \sin{\theta}) 
            & s_2 ( ( \beta + \delta \lambda_1) \cos{\theta} - (\beta \lambda_2 +  \delta) \sin{\theta}) \\
            s_1 ((\alpha + \gamma \lambda_1) \sin{\theta} + (\alpha \lambda_2  + \gamma) \cos{\theta})
            & s_2 ((\beta  + \delta \lambda_1) \sin{\theta} + (\beta \lambda_2 + \delta) \cos{\theta}) 
        \end{pmatrix} 
    \end{split}
\end{equation}

In order to add new pre-rasterization transformations, users can edit the function {\it render\_lt\_text} in the script {\it freetype\_text\_generator.py}. More specifically, this function contains 2 passes over the sequence of characters to synthesize (a sequence containing a single character is a special case), the first pass computes the bounding box, the second pass performs the actual rendering. In each pass, users can loop over the anchor points of each character and perform the required transformations accordingly in the font unit space~\cite{TrueType10:online, FreeType10:online, OpenType27:online}. Algorithm~\ref{alg:preelastic} shows an example.

\section{Post-rasterization transformations}

\begin{figure}[hb]
    \centering
    \includegraphics[width=\linewidth]{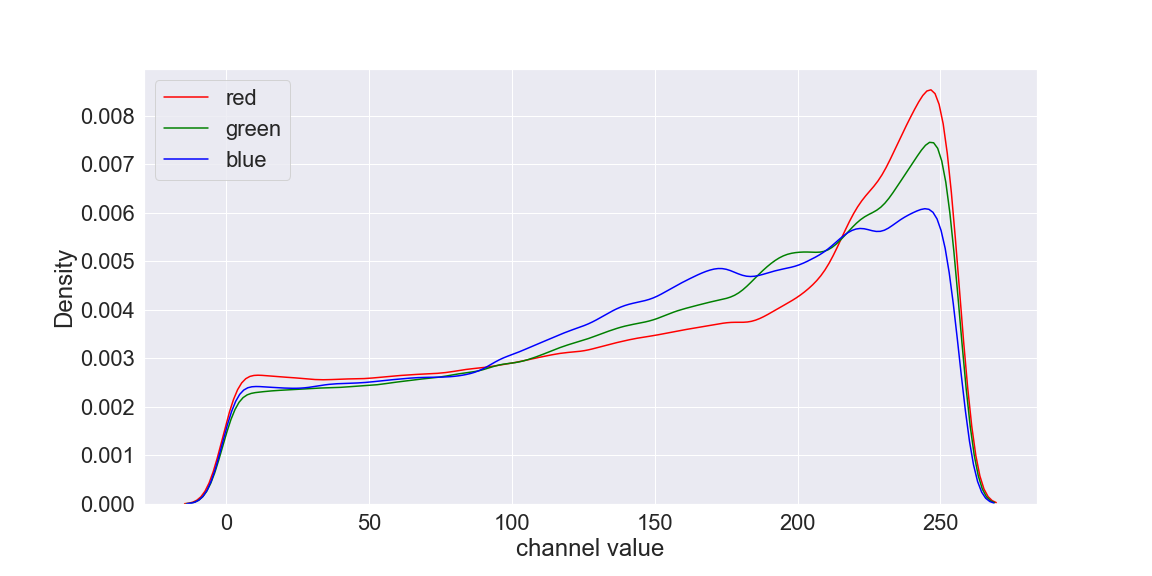}
    \caption{{\bf Kernel density estimation of the marginal color distribution.} Each curve is the estimated distribution of one color channel.}
    \label{fig:colordistribution}
\end{figure}

\begin{itemize}
    \item {\bf Translation} is performed, if any, when the foreground text is blended into the background.
    \item {\bf Perspective transformations} can be used to imitate the effect of different camera viewpoints. A perspective transformation is generally parameterized by a \(3\times 3\) matrix in homogeneous coordinates. The homogeneous matrix coefficients are computed from 4 pairs of 2D points in the two projection planes by solving a linear system. 
    \item {\bf Morphological image processing} is a set of operations on the shape of the character and they operate on binary images (foreground vs background). In total, 7 morphological transformations are available via OpenCV~\cite{opencv_library}: morphological erosion, morphological dilation, morphological opening, morphological closing, morphological gradient, Top Hat, Black Hat. 
    \begin{itemize}
        \item {\bf Morphological erosion} can be used to thin the stroke width in the post-rasterization phase. It erodes away the boundaries of foreground text and it can detach some previously connected strokes. The principle is to apply a 2D convolution, a pixel in the foreground text layer will be kept only if all the neighbor pixels are within the foreground area, otherwise it is eroded. The neighborhood is defined by a convolution kernel whose shape can be selected among rectangle, ellipse or cross-shaped. 
        \item {\bf Morphological dilation} can be used to thicken the stroke width in the post-rasterization phase and join detached strokes, which is the opposite of morphological erosion. A pixel will be put into the foreground if at least one neighbor pixel is within the foreground area. 
        \item {\bf Morphological opening} is the morphological erosion followed by morphological dilation. It can remove small pixel noises in the background, if any.
        \item {\bf Morphological closing} is the morphological dilation followed by the morphological erosion, which is the opposite of morphological opening. It can close small holes inside the foreground text, if any. 
        \item {\bf Morphological gradient} is the difference between morphological dilation and morphological erosion of the input image. It can render hollow text in the post-rasterization phase. 
        \item {\bf Top Hat} is the difference between the input image and the morphological opening of the input image. 
        \item {\bf Black Hat} is the difference between the morphological closing of the input image and the input image.
    \end{itemize}
    \item {\bf Gaussian blur} is implemented using scikit-image~\cite{walt_scikit-image_2014}. In the synthesis pipeline, Gaussian blur is usually applied before downsampling to avoid aliasing.
    \item {\bf Variation of contrast, brightness, color enhancement, sharpness} is implemented using Imgaug~\cite{imgaug}. 
    \item {\bf Elastic transformation (post-rasterization)}~\cite{simardBestPracticesConvolutional2003, imgaug} moves pixels locally around using displacement field. Depending on parameters, this transform can produce pixelated images or smooth deformation.
    \item {\bf Foreground filling} Foreground text can be filled either by uniform color or by natural image/texture. The sampling distribution (Figure~\ref{fig:colordistribution}) of random color is from \cite{kevinwuh7:online} (MIT License). When using random color for both foreground text and background, OmniPrint automatically ensures that foreground and background colors are visually distinguishable by thresholding the Delta E value (CIE2000). The computation of the Delta E value (CIE2000) is enabled by \cite{gtaylorp99:online} (BSD-3-Clause License). 
    \item {\bf Text outline} can be generated and filled either by uniform color or by natural image/texture.
    \item {\bf Background blending} can be done in two ways: (1) naively paste the foreground text onto the background while considering the mask; (2) Poisson Image Editing~\cite{10.1145/882262.882269} which ensures seamless blending, this is particularly useful in case of natural background. The implementation is from \cite{Gupta16}, which is under Apache License 2.0. Background can be filled by uniform color, natural image/texture or uniform color augmented with a random regular polygon. 
\end{itemize}

New post-rasterization transformations can be added to the image synthesis pipeline. For example, if one wants to add a transformation called {\it my\_transform}.

\begin{enumerate}
    \item Create a Python script called {\it my\_transform.py} under the directory {\it transforms}
    \item Implement the desired functionalities in {\it my\_transform.py}, which contains a function called {\it transform}. The first two positional parameters of the function {\it transform} should be the image and its corresponding mask (the mask is used for masking foreground text layer such that only the text itself will be pasted onto the background). The image is a RGB {\it PIL.Image.Image} object where text is black (0) and background is white (255). The mask is a grayscale {\it PIL.Image.Image} object where text is white (255) and background is black (0). In principle, the mask should undergo the same operations as the image while taking into account the difference in image mode and black/white convention. The function {\it transform} can, of course, accept other parameters, which is usually the case. The output of the function {\it transform} is a tuple of size 2: the first is the transformed image, the second is the transformed mask. 
    \item Edit the script {\it \_\_init\_\_.py} under the directory {\it transforms}, add one line: {\it from transforms.my\_transform import transform as my\_transform}
    \item Edit the script {\it data\_generator.py} to insert the implemented transform at appropriate location. For example, {\it img, mask = my\_transform(img, mask)}
    \item It is recommended to edit the argument parsing function of the entry script {\it run.py}, which allows specifying parameters of the newly implemented transformation via command line. It is also recommended to wrap {\it img, mask = my\_transform(img, mask)} under {\it data\_generator.py} by something like {\it if args.get(my\_transform) is not None:}, which allows to activate and deactivate the newly implemented transformation. 
\end{enumerate}

\section{Alphabets}

Here we present the character selection criteria:

\begin{itemize}
    \item For Latin script, we included basic uppercase and lowercase letters, all the variants in different European languages as well as the International Phonetic Alphabet. They are classified into basic Latin uppercase, basic Latin lowercase, Latin-1 Supplement, Latin Extended-A, Latin Extended-B, IPA letters and IPA for disordered speech and sinology, as defined in Unicode standard. 
    \item Chinese characters, also known as CJK Unified Ideographs, are numerous and their usage in real life are extremely imbalanced. In consequence, we only included Chinese characters from Table of General Standard Chinese Characters~\cite{TableofG17:online}. These Chinese characters are divided into three levels containing 3500, 3000 and 1605 characters respectively. Characters in group 1 and 2 (the first 6500) are designated as common. Different from other writing systems, the distinction between simplified Chinese characters, traditional Chinese characters, Japanese Kanji and Korean Hanja is only handled by fonts in principle, because many of them share the same code points. The only way to distinguish them is the fonts' rendering. Generally, the fonts that were designed for simplified Chinese characters should never be used when rendering traditional Chinese text or Japanese text, and vice versa. Otherwise, it can be unintelligible or be unacceptable for native speakers. To avoid this overhead, we only aim to render simplified Chinese characters. 
    \item For Japanese, all of Hiragana and Katakana are included. Note that each letter of these two scripts appears twice in the Unicode standard, one corresponds to the normal-sized version, the other is the smaller version. We only included the normal-sized versions. 
    \item For Korean, there are up to 11172 unique syllabic blocks, we only included 2350 syllabic blocks which are assumed to be commonly used. 
    \item All letters of Cyrillic script are not included. Only modern Russian alphabet is included, which consists of 66 upper case and lower case letters. 
    \item Writing systems like Abjad (Arabic, Hebrew, etc.) and Abugida (Thai, Lao, Tibetan, Devanagari, Bengali, etc.) are only partly included. Typically, we only included consonants, independent vowels and digits of these languages. For these scripts (Khmer, Balinese, Bengali, Devanagari, Gujarati, Myanmar, Oriya, Sinhala, Tamil, Telugu, Tibetan, Thai and Lao.), dependent vowel signs were excluded, independent vowels were included if there are any. 
    \item Even though the Mongolian script has been adapted to write languages such as Oirat and Manchu, we only included basic Mongolian letters and Mongolian digits.
    \item For the Arabic script, we only included the 29 Arabic letters. For the Hebrew script, we only included the 27 Hebrew letters. 
    \item All of the Ethiopic syllables available in the Unicode standard are included. 
    \item Common punctuations and symbols, ASCII digits, some musical symbols and some mathematical operators are also included. However, neither of the collected fonts fully support these musical symbols. 
\end{itemize}

\section{Accessibility}


The NeurIPS foundation shall not bear any responsibility. The diffusion of the code and data will be done by the authors, who will be responsible for maintaining them and resolving any dispute. 
\begin{itemize}
    \item The code of the OmniPrint data synthesizer will be made available on Github under an open source MIT license \href{https://opensource.org/licenses/MIT}{https://opensource.org/licenses/MIT}. A specific guarantee disclaimer is associated with the license:
\texttt{THE SOFTWARE IS PROVIDED "AS IS", WITHOUT WARRANTY OF ANY KIND, EXPRESS OR IMPLIED, INCLUDING BUT NOT LIMITED TO THE WARRANTIES OF MERCHANTABILITY, FITNESS FOR A PARTICULAR PURPOSE AND NONINFRINGEMENT. IN NO EVENT SHALL THE AUTHORS OR COPYRIGHT HOLDERS BE LIABLE FOR ANY CLAIM, DAMAGES OR OTHER LIABILITY, WHETHER IN AN ACTION OF CONTRACT, TORT OR OTHERWISE, ARISING FROM, OUT OF OR IN CONNECTION WITH THE SOFTWARE OR THE USE OR OTHER DEALINGS IN THE SOFTWARE.}
    \item The datasets OmniPrint-meta[1-5] will be distributed via the UCI repository and/or Kaggle datasets. They will be licensed under a Creative Commons license CC BY 4.0 \href{https://creativecommons.org/licenses/by/4.0/}{https://creativecommons.org/licenses/by/4.0/}. This comes with the following guarantee disclaimer: \texttt{Unless otherwise separately undertaken by the Licensor, to the extent possible, the Licensor offers the Licensed Material as-is and as-available, and makes no representations or warranties of any kind concerning the Licensed Material, whether express, implied, statutory, or other. This includes, without limitation, warranties of title, merchantability, fitness for a particular purpose, non-infringement, absence of latent or other defects, accuracy, or the presence or absence of errors, whether or not known or discoverable. Where disclaimers of warranties are not allowed in full or in part, this disclaimer may not apply to You. To the extent possible, in no event will the Licensor be liable to You on any legal theory (including, without limitation, negligence) or otherwise for any direct, special, indirect, incidental, consequential, punitive, exemplary, or other losses, costs, expenses, or damages arising out of this Public License or use of the Licensed Material, even if the Licensor has been advised of the possibility of such losses, costs, expenses, or damages. Where a limitation of liability is not allowed in full or in part, this limitation may not apply to You.}
\end{itemize}

In general, modern digital fonts are protected under their own licenses, we do not provide any warranty for this. Some fonts cannot be redistributed or modified. However, the users are free to collect or make their own fonts.




The code\footnote{\href{https://github.com/SunHaozhe/OmniPrint}{https://github.com/SunHaozhe/OmniPrint}} and datasets\footnote{\href{https://github.com/SunHaozhe/OmniPrint-datasets}{https://github.com/SunHaozhe/OmniPrint-datasets}} have been publicly released after the NeurIPS 2021 meta-learning challenge. The hosting platform of the datasets OmniPrint-meta[1-5] is \href{https://www.kaggle.com/datasets}{Kaggle Datasets}, DOI for datasets is 10.34740/kaggle/dsv/2763401, metadata is accessible on the dataset hosting page. 

\href{https://www.kaggle.com/datasets}{Kaggle Datasets} make data available for an unlimited time period. The authors will verify that the data are properly accessible for at least three years and change venue in case of a problem. Likewise GitHub has no time limitations in terms of code hosting.  The authors will maintain the code and address issues for at least three years. Users will be encouraged to post GitHub issues in case of problems and/or make pull requests.

Any information and updates regarding to the {\bf release} and necessary {\bf maintenance} will be communicated via the README of \href{https://github.com/SunHaozhe/OmniPrint-datasets}{https://github.com/SunHaozhe/OmniPrint-datasets}. 




Each dataset synthesized by OmniPrint shares the same folder structure. It contains two subfolders, the subfolder {\it data} contains the images in png format, the subfolder {\it label} contains a csv file, called {\it raw\_labels.csv}, which stores the label (character class) as well as all the metadata of each image instance. The columns of {\it raw\_labels.csv} may vary depending on involved transformations, the common columns include {\it image\_name} which specifies which image instance this record is about, {\it text} which contains the rendered character to synthesize, {\it unicode\_code\_point} contains the Unicode code point (integer) of the character to synthesize, {\it font\_file} which indicates the involved digital font, {\it background} which specifies which type of background is being used, {\it font\_weight} which specifies the stroke width, {\it margin\_bottom, margin\_left, margin\_right, margin\_top} which indicate the proportion of each margin in the image and facilitate the construction of bounding boxes, {\it family\_name, style\_name} which show the family and font style to which the digital font belongs, etc. 

The user manual of the data synthesizer OmniPrint and an example dataloader for the datasets OmniPrint-meta[1-5] are provided with the \href{https://github.com/SunHaozhe/OmniPrint}{code}. 


\section{Experimental details of domain adaptation}

This section provides the experimental details of Section~4.4 of the main paper.

\subsection{Unsupervised domain adaptation methods}
\label{sec:unsuperviseddamethods}

The 5 unsupervised domain adaptation algorithms are DAN~\cite{longLearningTransferableFeatures2015, tzengDeepDomainConfusion2014}, DANN~\cite{ganinUnsupervisedDomainAdaptation2015}, DeepCoral~\cite{sunDeepCORALCorrelation2016}, DAAN~\cite{yuTransferLearningDynamic2019} and DSAN~\cite{zhuDeepSubdomainAdaptation2021}. The implementation is from DeepDA~\cite{deepdaJindong} which is under MIT License. 

\subsection{Hyperparameters and compute resources}

For each combination of task and algorithm, we run 10 epochs with 8 random seeds to get the confidence interval. The 8 random seeds were fixed in advance. The backbone neural network is Resnet50~\cite{heDeepResidualLearning2015}. The model is optimized using SGD~\cite{2016arXiv160904747R} with \(10^{-3}\) as the learning rate. No hyperparameter optimization was performed. The other experimental details are provided with the code at \href{https://github.com/SunHaozhe/transferlearning}{https://github.com/SunHaozhe/transferlearning}. The experiments were run on an internal cluster with Tesla V100-PCIE-32GB, Tesla V100-SXM2-32GB. The total amount of computation time is about 182 hours. 

The results are available in Table~\ref{tab:transfer}. 

\begin{table}
  \caption{{\bf Unsupervised domain adaptation results} on OmniPrint-metaX-31. \(\text{meta}A\rightarrow \text{meta}B\) means the source domain is OmniPrint-meta\(A\), the target domain is OmniPrint-meta\(B\), where \(A, B \in {3, 4, 5}\). The \(95\%\) confidence intervals are computed with 8 random seeds.}
  \label{tab:transfer}
  \centering
  \begin{adjustbox}{width=\linewidth}
  \begin{tabular}{lllllll}
    \toprule
    & meta3\(\rightarrow\)meta4 & meta4\(\rightarrow\)meta3 & meta3\(\rightarrow\)meta5 & meta5\(\rightarrow\)meta3 & meta4\(\rightarrow\)meta5 & meta5\(\rightarrow\)meta4  \\
    \midrule
    DAN~\cite{longLearningTransferableFeatures2015, tzengDeepDomainConfusion2014} & 18.0 \(\pm\) 2.4  &    3.2 \(\pm\) 0.0 &    25.8 \(\pm\) 1.7  &    3.5 \(\pm\) 0.3   & 10.1 \(\pm\) 16.0  &  10.7 \(\pm\) 16.5 \\
    DANN~\cite{ganinUnsupervisedDomainAdaptation2015} & 72.2 \(\pm\) 2.8   &  96.8 \(\pm\) 0.5    & 65.6 \(\pm\) 2.9   &  82.2 \(\pm\) 2.7   &  79.8 \(\pm\) 1.2  &   81.5 \(\pm\) 2.1 \\
    DeepCoral~\cite{sunDeepCORALCorrelation2016} & 22.9 \(\pm\) 2.5  &   84.6 \(\pm\) 1.5  &   28.6 \(\pm\) 1.7  &   69.6 \(\pm\) 2.5  &   57.0 \(\pm\) 1.3  &   60.2 \(\pm\) 1.0 \\
    DAAN~\cite{yuTransferLearningDynamic2019} & 22.3 \(\pm\) 1.8&     84.5 \(\pm\) 2.1   &  25.1 \(\pm\) 1.7  &   59.9 \(\pm\) 5.9   &  50.9 \(\pm\) 1.5 &    53.3 \(\pm\) 2.3 \\
    DSAN~\cite{zhuDeepSubdomainAdaptation2021} & \(\mathbf{79.3 \pm 2.3}\) &    \(\mathbf{96.9 \pm 0.3}\)  &   \(\mathbf{66.4\pm 2.5}\)   &  \(\mathbf{93.5 \pm 0.8}\)   &  \(\mathbf{80.5 \pm 1.0}\) &   \(\mathbf{82.8 \pm 1.9}\) \\
    \cmidrule{2-7}
    Average & 42.9     &       73.2      &      42.3      &     61.7      &     55.7    &        57.7 \\
    Median & 22.9   &         84.6    &        28.6    &        69.6           &   57.0      &      60.2 \\
  \bottomrule
  \end{tabular}
  \end{adjustbox}
\end{table}

\subsection{Unsupervised domain adaptation from Fake-MNIST to MNIST}

We used OmniPrint to generate a dataset, called Fake-MNIST, which is similar to MNIST~\cite{lecun2010mnist} and performed the unsupervised domain adaptation (the 5 DeepDA methods~\cite{deepdaJindong}, see Appendix~\ref{sec:unsuperviseddamethods}) from Fake-MNIST to MNIST. 

\begin{figure}[htb]
    \centering
    \includegraphics[width=\linewidth]{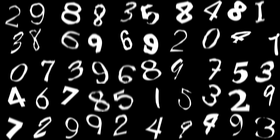}
    \caption{{\bf Example images from Fake-MNIST.} Random pre-rasterization elastic transformation, horizontal shear, rotation and translation were used.}
    \label{fig:fake_mnist}
\end{figure}

Only the test set of MNIST is involved in this experiment, which consists of 10000 images for the 10 digits. Fake-MNIST contains 3000 white-on-black character images for each of the 10 digits. Random pre-rasterization elastic transformation, horizontal shear, rotation and translation were used to synthesize Fake-MNIST. Figure~\ref{fig:fake_mnist} shows some example images from Fake-MNIST.

While the synthesis parameters of Fake-MNIST were not optimized, the performance of the 5 unsupervised domain adaptation methods (Table~\ref{tab:transfermnist}) ranges from 92 to 98\% accuracy, which is very honorable (current supervised learning results on MNIST are over 99\%).

\begin{table}
  \caption{{\bf Unsupervised domain adaptation from Fake-MNIST to MNIST}. 95\% confidence intervals are computed with 27 random seeds.}
  \label{tab:transfermnist}
  \centering
  {\footnotesize 
  \begin{adjustbox}{width=\linewidth}
  \begin{tabular}{llllllll}
    \toprule
    & DAN    &     DANN   & DeepCoral  &       DAAN      &   DSAN & Average & Median \\ 
    \midrule
    Fake-MNIST \(\rightarrow\) MNIST    &  94.8 \(\pm\) 0.1 &  98.0 \(\pm\) 0.1 &  92.4 \(\pm\) 0.2 & 93.3 \(\pm\) 0.2 & \(\mathbf{98.2 \pm 0.1}\) &  95.34  & 94.8  \\
  \bottomrule
  \end{tabular}
  \end{adjustbox}
  }
\end{table}

\section{Experimental details of few-shot learning experiments with metadata-based episodes}

This section provides the experimental details of Section~4.2 of the main paper.

\subsection{Metadata-based episode generation algorithm}

The metadata-based episode generation algorithm is illustrated in Algorithm~\ref{alg:z_episodes}.

\begin{algorithm}
\caption{Metadata-based few-shot learning episode generation.}
\label{alg:z_episodes}
\SetAlgoLined
\DontPrintSemicolon
\KwIn{Number of support images \(S\), number of query images \(Q\)}
\tcp*[l]{Assuming that metadata consists of real numbers.} 
\For{\textup{each episode}}{
    Randomly sample \(N\) classes \(c_1, c_2, ..., c_N\) \;
    \For{\textup{each class} \(c_n\)}{
        Find all examples \(E_{c_n} = \{e_1, e_2, ...\}\) of class \(c_n\), the metadata \(m_i\) of each example \(e_i \in E_{c_n}\) is a real-valued vector. \;
        Compute the bounding box \(B_{c_n}\) of the metadata vectors \(m_i\). \;
        Randomly sample a centroid \(D\) within \(B_{c_n}\). \;
        Select the \((S+Q)\) nearest neighbors \(M = \{m_x, m_y, ..., m_{(S+Q)}\}\) from all the metadata vectors \(m_1, m_2, ...\) \;
        An example \(e_i\) is selected to be part of the episode if and only if \(m_i \in M\), all the selected examples form the set \(\hat{E}_{c_n, D}\) \;
        Randomly draw \(S\) examples from \(\hat{E}_{c_n, D}\) to form the support set, the remaining examples serve as the query set. 
    }  
}
\end{algorithm}

\subsection{Data, hyperparameters and compute resources}

The experiments used the same hyperparameters as Appendix~\ref{sec:experimentaldetails}. The same character split was used (900 characters for meta-train, 149 characters for meta-validation, 360 characters for meta-test). The experiments were trained for 300 epochs, where each epoch is defined to be 6 batches of episodes, each batch contains 32 episodes. During meta-training, the model checkpoints were evaluated on meta-validation episodes every 5 epochs. Only the checkpoint having the highest accuracy on meta-validation episodes during training is selected to be tested on meta-test episodes. The backbone neural network is the concatenation of three modules of Convolution-BatchNorm-Relu-Maxpool. The reported accuracy and 95\% confidence intervals were computed with 5 random seeds. 

The experiments were run on an internal cluster, the involved GPUs are GeForce RTX 2080 Ti, Tesla V100-PCIE-32GB and Tesla V100-SXM2-32GB. The total amount of computation time is about 164 hours.

\section{Experimental details of the investigation of the influence of the number of meta-training episodes}

This section provides the experimental details of Section~4.3 of the main paper.

\subsection{Data}

For this experiment, we generated a larger version of OmniPrint-meta3. It has the same synthesis parameters as OmniPrint-meta3 but has 200 images per class (OmniPrint-meta3 has 20 images per class). 

\subsection{Hyperparameters and compute resources}

The experiments used the same hyperparameters as Appendix~\ref{sec:experimentaldetails}. The same character split was used (900 characters for meta-train, 149 characters for meta-validation, 360 characters for meta-test). During meta-training, the model checkpoints were evaluated on meta-validation episodes every 960 episodes and at the end of meta-training. Only the checkpoint having the highest accuracy on meta-validation episodes during training is selected to be tested on meta-test episodes. The backbone neural network is the concatenation of three modules of Convolution-BatchNorm-Relu-Maxpool. The reported accuracy and 95\% confidence intervals were computed with 5 random seeds. 

The experiments were run on an internal cluster, the involved GPUs are GeForce RTX 2080 Ti, Tesla V100-PCIE-32GB. The total amount of computation time is about 80 hours.

\section{Experimental details of the regression task}

This section provides the experimental details of Section~4.5 of the main paper.

\subsection{Data}

\begin{figure}[h]
    \centering
    \begin{minipage}{0.46\linewidth}
        \centering
        \includegraphics[width=\linewidth]{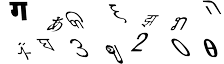}
        \caption{{\bf Shear dataset.} Horizontal shear parameter ranges from -0.8 to 0.8. Rotation and perspective transformations are not used.}
        \label{fig:shear_x_dataset}
    \end{minipage}%
    \hspace{10mm}
    \begin{minipage}{0.46\linewidth}
        \centering
        \includegraphics[width=\linewidth]{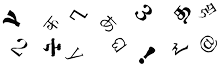}
        \caption{{\bf Rotation dataset.} Rotation angle ranges from -60 degrees to 60 degrees. Horizontal shear and perspective transformations are not used.}
        \label{fig:rotation_dataset}
    \end{minipage}
\end{figure}

We generated two large datasets which are slightly easier than OmniPrint-meta3. Both datasets contain black-on-white characters (1409 characters with 200 images each). The first dataset has horizontal shear (horizontal shear parameter ranges from -0.8 to 0.8) but not rotation, the second dataset has rotation (rotation ranges from -60 degrees to 60 degrees) but not horizontal shear. Perspective transformations are not used. Some sample images are shown in Figure~\ref{fig:shear_x_dataset} and Figure~\ref{fig:rotation_dataset}. Each of the two generated datasets have 281800 images in total. 20\% of the images (56360) were used for validation, 20\% of the images (56360) were used for test. The remaining 169080 images were used for training.

\subsection{Hyperparameters and compute resources}

We tested two neural networks. The first one, referred to as "small", is the concatenation of three modules of Convolution-BatchNorm-Relu-Maxpool, followed by a fully-connected layer within a scalar output. It contains 76097 trainable parameters. The second one is Resnet18~\cite{heDeepResidualLearning2015} pretrained on ImageNet~\cite{2014arXiv1409.0575R}. We only train the last convolution layer and fully-connected layer of Resnet18~\cite{heDeepResidualLearning2015}, it thus has 2360833 trainable parameters. The neural networks were optimized with MSE loss for 30 epochs using SGD~\cite{2016arXiv160904747R}, the initial learning rate was \(10^{-3}\), which is reduced by a factor of 10 when the validation loss has stopped decreasing for 5 epochs. The weight decay was \(10^{-4}\). The momentum was \(0.9\). Only the model having the highest accuracy on validation data during training is selected to be tested on test data. The \(95\%\) confidence intervals are computed with 3 random seeds.

The experiments were run on an internal cluster with GeForce RTX 2080 Ti. The total amount of computation time is about 48 hours.

The detailed results are reported in Table~\ref{tab:regressiontable}.

\begin{table}
  \caption{{\bf Regression results}. The reported metric is the coefficient of determination \(R^2\). \(1.69\times 10^{2}\), \(1.69\times 10^{3}\), \(1.69\times 10^{4}\) and \(1.69\times 10^{5}\) are the number of training images. 95\% confidence intervals are computed with 3 random seeds.}
  \label{tab:regressiontable}
  \centering
  {\footnotesize 
  \begin{tabular}{lllllll}
    \toprule
    & Backbone & \(1.69\times 10^{2}\)     &    \(1.69\times 10^{3}\)     &   \(1.69\times 10^{4}\)    &   \(1.69\times 10^{5}\)  \\
    \midrule
    \multirow{2}{*}{Horizontal shear}& small & 0.3 \(\pm\) 0.2 &  0.6 \(\pm\) 0.0 & 0.8 \(\pm\) 0.1 & 0.9 \(\pm\) 0.0 \\
    & resnet18 & -0.1 \(\pm\) 0.2 & 0.6 \(\pm\) 0.0 & 0.8 \(\pm\) 0.0 & 0.9 \(\pm\) 0.0 \\
    \cmidrule{2-6}
    \multirow{2}{*}{Rotation}& small & -25.3 \(\pm\) 50.4 & -1.8 \(\pm\) 0.4 & -0.8 \(\pm\) 1.7 & 0.3 \(\pm\) 0.2 \\
    & resnet18 & -1002.0 \(\pm\) 3164.1 & -0.9 \(\pm\) 0.2 &  0.1 \(\pm\) 0.1 & 0.5 \(\pm\) 0.0 \\
  \bottomrule
  \end{tabular}
  }
\end{table}


\FloatBarrier

\medskip

{
\small
\bibliography{supplement}
}